%% file: PaperCameraReady.tex
\documentclass[10pt,twocolumn,letterpaper]{article}

\usepackage[pagenumbers]{wacv} %

\usepackage[accsupp]{axessibility}
\usepackage{graphicx}
\usepackage{grffile}
\usepackage{amsmath}
\usepackage{amssymb}
\usepackage{booktabs}
\usepackage{multicol}
\usepackage{enumerate}
\usepackage[table, usenames, dvipsnames]{xcolor}
\usepackage{float}
\usepackage{siunitx} 
\usepackage[colorinlistoftodos,prependcaption,textsize=footnotesize]{todonotes}

\input{shortcuts}

\input{all_figures}

\input{all_tables}
\input{preamble.tex}

\usepackage[pagebackref,breaklinks,colorlinks]{hyperref}

\usepackage[capitalize]{cleveref}
\crefname{section}{Sec.}{Secs.}
\Crefname{section}{Section}{Sections}
\Crefname{table}{Table}{Tables}
\crefname{table}{Tab.}{Tabs.}

\def\wacvPaperID{1163} %
\def\confName{WACV}
\def\confYear{2024}

\begin{document}

\input{title}
\title{\thetitle}

\author{
Sebastian Koch$^{1,2,3}$\qquad Pedro Hermosilla$^4$\qquad Narunas Vaskevicius$^{1,2}$\qquad \\Mirco Colosi$^2$\qquad Timo Ropinski$^3$ \vspace{0.05cm}\\ 
{\small $^1$\text{Bosch Center for Artificial Intelligence}\quad $^2$Robert Bosch Corporate Research}\\ {\small $^3$\text{University of Ulm} \quad $^4$ TU Wien}
\\{\small\href{https://kochsebastian.com/sgrec3d}{kochsebastian.com/sgrec3d}}
}
\maketitle

\input{content/abstract.tex}
\input{content/introduction.tex}
\input{content/related_work.tex}

\input{content/method.tex}

\input{content/experiments_new.tex}

\input{content/conclusion.tex}

\newpage
\vspace{0.5em}\noindent\textbf{Acknowledgement}
 This work was partly supported by the EU Horizon 2020 research and innovation program under grant agreement No. 101017274 (DARKO).
{\small
\bibliographystyle{ieee_fullname}
\bibliography{egbib}
}
\newpage

\input{content/supp.tex}
\end{document}

%% file: shortcuts.tex
\makeatletter
\DeclareRobustCommand\onedot{\futurelet\@let@token\@onedot}
\def\@onedot{\ifx\@let@token.\else.\null\fi\xspace}

\def\etc{etc\onedot}

\def\etal{et~al\onedot}

\makeatother

\newcommand{\boldparagraph}[1]{\vspace{0.5em}\noindent{\bf #1.}}

\renewcommand{\paragraph}[1]{\boldparagraph{#1}}

\definecolor{darkgreen}{rgb}{0,0.7,0}
\definecolor{newyellow}{rgb}{1,0.8,0.05}
\definecolor{newgreen}{rgb}{0.2,0.8,0.2}

\newcolumntype{P}[1]{>{\centering\arraybackslash}m{#1}}

%% file: all_figures.tex
\newcommand{\figteaser}{
\begin{figure}[t]
  \centering
   \includegraphics[width=\linewidth]{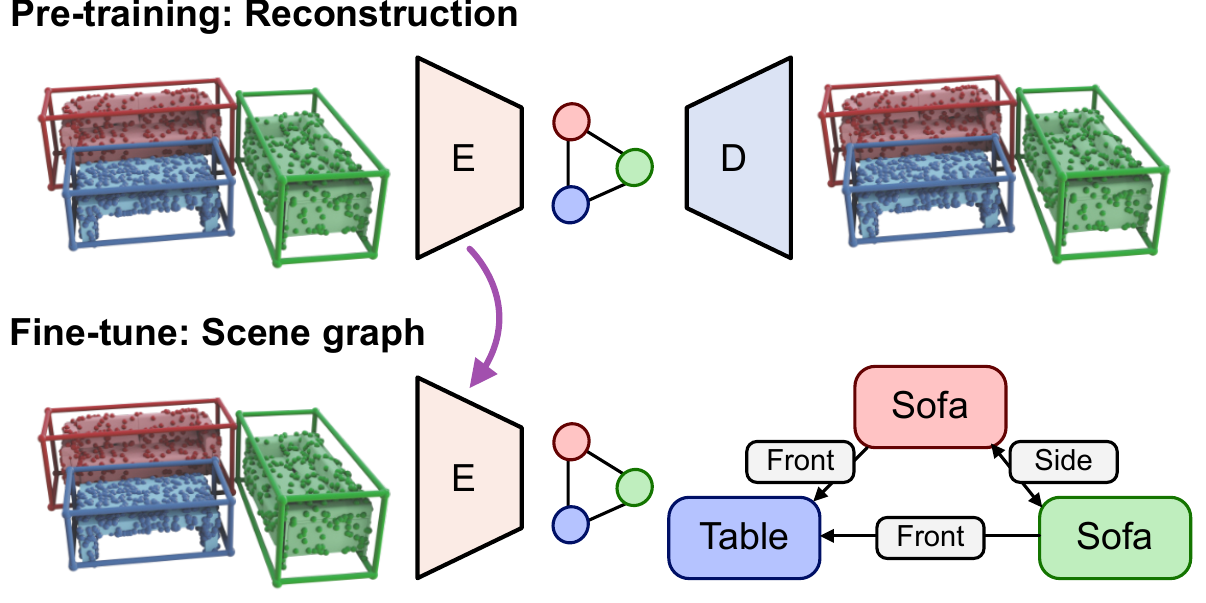}
   \caption{\textbf{\ours Overview.} \ours~exploits autoencoder-based pre-training to build a 3D scene graph latent space through a reconstruction loss. The resulting encoder can, later on, be fine-tuned for the downstream 3D scene graph prediction.}
   \label{fig:teaser}
   \vspace{-0.5cm}
\end{figure}
}

\newcommand{\figmainmethod}{
\begin{figure*}[t]
  \centering
  \includegraphics[width=\textwidth]{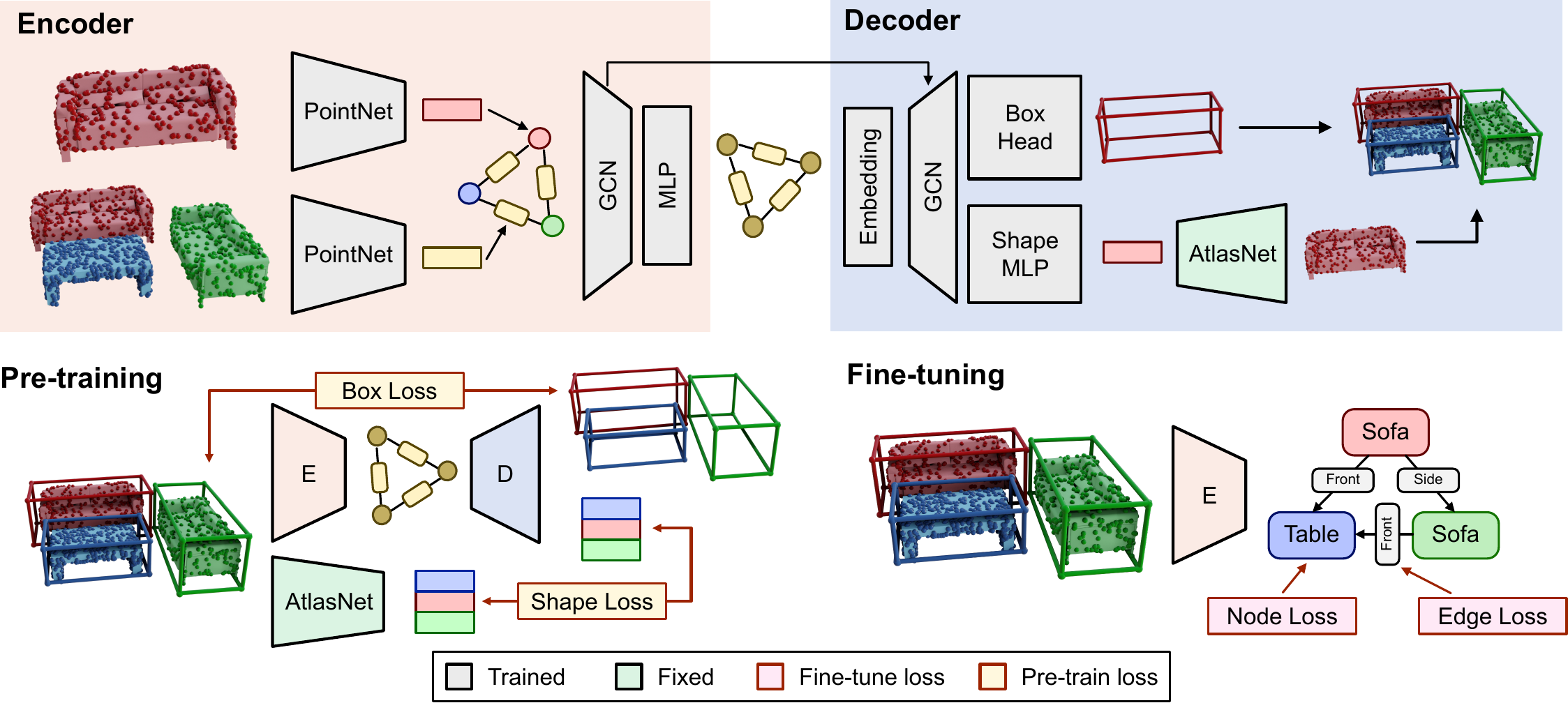}
  \caption{
  \textbf{\ours architecture overview.}
  \new{
  \ours utilizes an encoder-decoder structure for pre-training (bottom-left) using a reconstruction loss by reconstructing the bounding boxes and shape encodings of objects with supervision from a pre-trained AtlasNet. 
  The encoder (top-left) generates object and edge embeddings from the input point cloud into a latent graph. The decoder (top-right) reconstructs the input point cloud from the graph bottleneck. During fine-tuning (bottom-right), the decoder is discarded and the encoder is fine-tuned to predict the node and edge classes.
  }
  }
  \label{fig:short}
  \vspace*{-.5cm}
\end{figure*} 
}

\newcommand{\figcomparison}{
\begin{figure*}[t]
  \centering
  \includegraphics[width=0.97\linewidth]{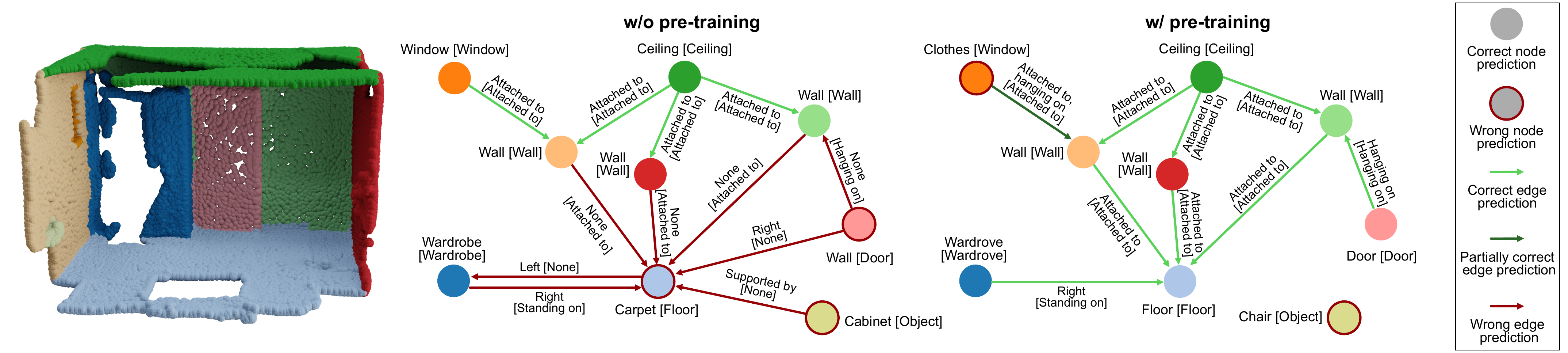}
  \caption{\textbf{Qualitative evaluation of the effects of pre-training on 3DSSG.} A qualitative comparison of a scene graph predicted for a 3D scene (\emph{left}), with no pre-training (\emph{middle}) and with pre-training~(\emph{right}). We visualize the top-1 object class prediction for each node and the predicates with a probability greater than 0.5 for each edge. Ground truth labels are shown in square brackets. \ours~results in 7/9 correct nodes and 9/9 (partially) correct predicate predictions, while the baseline only reaches 5/9 and 4/9 respectively, despite predicting many false positive predicates.}
  \label{fig:graphs_example_compare}
\end{figure*}
}

\newcommand{\figexamples}{
\begin{figure*}[t]
  \centering
  \includegraphics[width=0.97\textwidth, keepaspectratio]{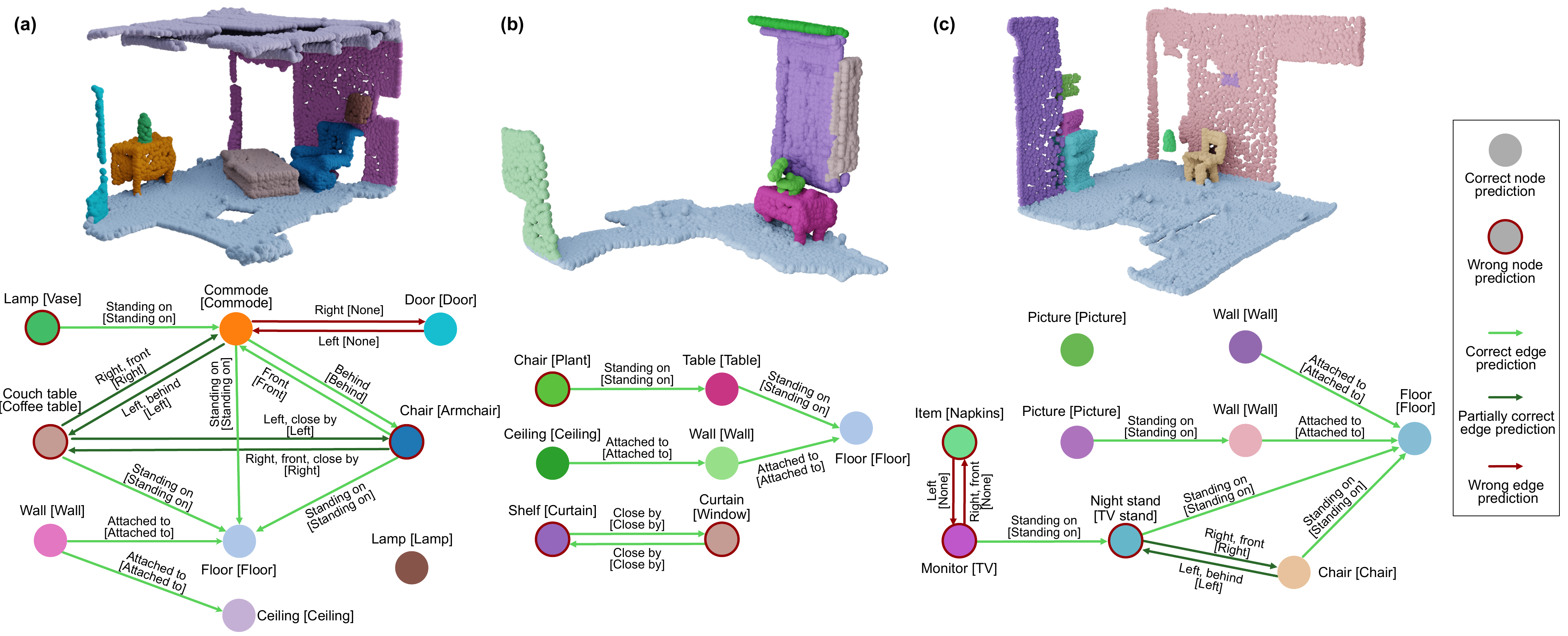}
  \caption{\textbf{3D scene graph visualizations for 3DSSG scene splits.} We visualize the top-1 object class prediction for each node and the predicates with a probability greater than 0.5 for each edge. Ground truth labels are shown in square brackets.}
  \label{fig:scene_graph_visualisation}
    \vspace*{-.5cm}
\end{figure*}
}

\newcommand{\figreconstructions}{\begin{figure*}[t]
  \centering
  \includegraphics[width=0.97\linewidth,keepaspectratio]{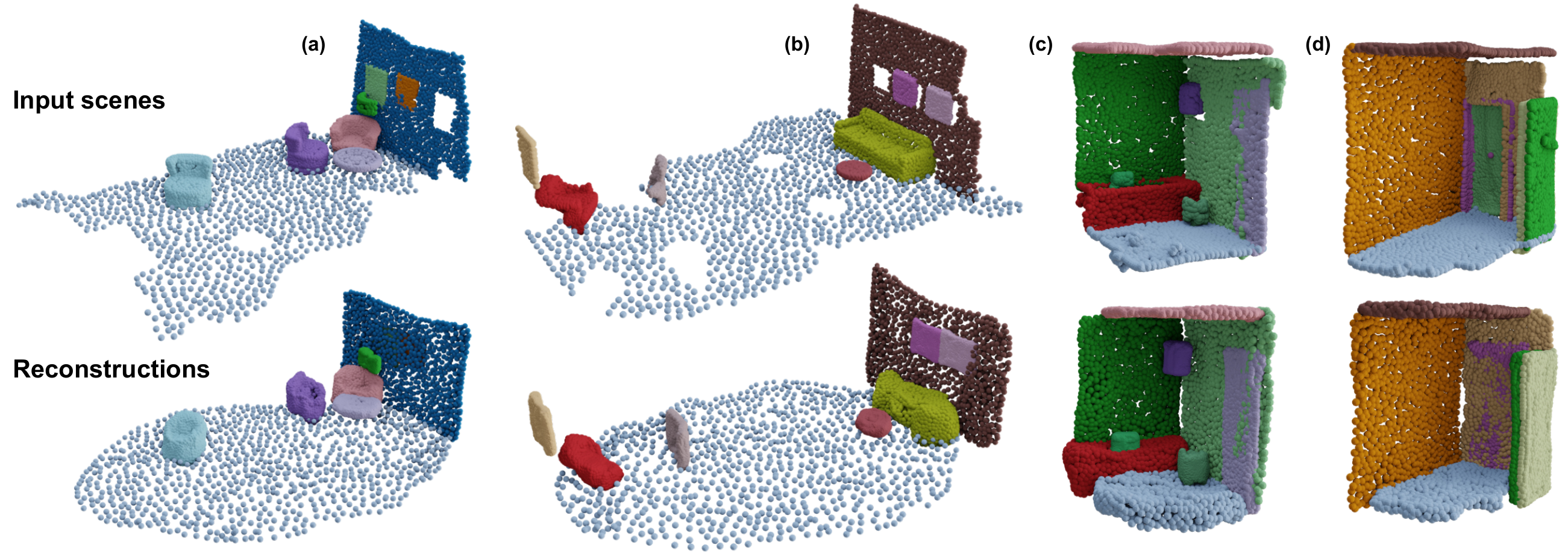}
  \caption{\textbf{Scene reconstruction on 3DSSG.} Qualitative results of the reconstructions for four different input scenes. While the reconstruction does not represent the input scenes perfectly, it is faithful to the input scenes, as object locations are well-preserved. Furthermore, the shapes look similar and the reconstruction quality is stable.}
  \label{fig:scene_reconstructions}
    \vspace*{-.4cm}
\end{figure*}
}

%% file: all_tables.tex
\newcommand{\tabsmall}[1]{{\scriptsize #1}}

\newcommand{\tabmaineval}{
\begin{table}[t]
\tabcolsep=1.2mm
\centering
\small
\begin{tabular}{@{}lcccccc@{}}
\toprule
       & \multicolumn{2}{c}{Object} & \multicolumn{2}{c}{Predicate} & \multicolumn{2}{c}{Relationship} \\
       \cmidrule(lr){2-3}\cmidrule(lr){4-5}\cmidrule(lr){6-7}
      Method & R@5                  & R@10                 & R@3                 & R@5                & R@50                 & R@100                \\
      \midrule
3D + MSDN \cite{Li_2017_ICCV} & 0.61 & 0.72 & 0.86 & 0.94 & 0.47 & 0.53\\
3D + KERN \cite{Chen_2019_CVPR} & 0.67 & 0.77 & 0.83 & 0.96 & 0.51 & 0.58\\ 
3D + BGNN \cite{Li_2021_CVPR} & 0.71 & 0.82 & 0.87 & 0.94 & 0.55 & 0.60\\
SGGPoint \cite{Zhang_2021_CVPR} & 0.28 & 0.36 & 0.68 & 0.87 & 0.08 & 0.10\\
3DSSG \cite{Wald_2020_CVPR} & 0.68                 & 0.78                 & 0.89                & 0.93               & 0.40                 & 0.66                 \\ 
Liu \etal \cite{Liu_2022_TVCG} & 0.74 & 0.83 & 0.90 & 0.96 & 0.62 & 0.68\\
SGFN \cite{Wu_2021_CVPR} & 0.70 & 0.80 & \textbf{0.97} & \textbf{0.99} & 0.85 & 0.87 \\
Ours & \textbf{0.80}                 & \textbf{0.87}                 & \textbf{0.97}                & \textbf{0.99}               & \textbf{0.89}                 & \textbf{0.91}                \\ 
\bottomrule
\end{tabular}
\caption{\textbf{3D scene graph prediction on 3DSSG.}  Experimental results for 3D scene graph prediction on 3DSSG. We report the top-k recall values for object classification, predicate prediction as well as relationship prediction. For a fair comparison, all works use ground-truth class-agnostic instance segmentation.}
\label{tab:main_evaluation}
\vspace{-0.5cm}
\end{table}
}

\newcommand{\tabreconeval}{
\begin{table}[t]
\centering
\tabcolsep=4pt
\small
\begin{tabular}{@{}lccc@{}}
\toprule
Relationship & \multicolumn{1}{c}{Rule} &  GraphTo3D & Ours \\\midrule
left of & \multicolumn{1}{c}{\tabsmall{$c_{x,i}<c_{x,j}$ and $iou(b_i,b_j)<0.5$}} &  0.85 & 0.92\\
right of & \multicolumn{1}{c}{\tabsmall{$c_{x,i}>c_{x,j}$ and $iou(b_i,b_j)<0.5$}} &  0.85 & 0.92\\
front of & \multicolumn{1}{c}{\tabsmall{$c_{y,i}<c_{y,j}$ and $iou(b_i,b_j)<0.5$}} &  0.79 & 0.90\\ 
behind of & \multicolumn{1}{c}{\tabsmall{$c_{y,i}>c_{y,j}$ and $iou(b_i,b_j)<0.5$}} &  0.79 & 0.90\\
higher than & \tabsmall{$h_i+c_{z,i}/2>h_j+c_{z,j}/2$} & 0.96 & 0.96\\
lower than & \tabsmall{$h_i+c_{z,i}/2<h_j+c_{z,j}/2$} & 0.96 & 0.96 \\
smaller than & \tabsmall{$w_il_ih_i<w_jl_jh_j$} &  0.98 & 0.96\\
bigger than & \tabsmall{$w_il_ih_i>w_jl_jh_j$} & 0.98 & 0.96 \\
same as & \tabsmall{$iou(b_i,b_j)>0.5$} &  1.00 & 1.00\\ \midrule
average & & 0.90 & 0.94\\ \bottomrule
\end{tabular}
\caption{\textbf{Rule-based scene generation verification.} A simple rule-based verification of the scene generations from \ours, for a subset of the predicates in the 3DSSG dataset.}
\label{tab:rule_verify}
\vspace{-0.7cm}
\end{table}
}

\newcommand{\tabdataseteval}{

\begin{table}[t]
\tabcolsep=0.3mm
\centering
\small
\begin{tabular}{@{}lcccccc@{}}
\toprule
       & \multicolumn{2}{c}{Object} & \multicolumn{2}{c}{Predicate} & \multicolumn{2}{c}{Relationship} \\
       \cmidrule(lr){2-3}\cmidrule(lr){4-5}\cmidrule(lr){6-7}
      Dataset & R@5                  & R@10                 & R@3                 & R@5                & R@50                 & R@100                \\   
      \midrule
3DSSG &  0.75                 & 0.83                & 0.96            & \textbf{0.99}              & 0.88                 & 0.89                \\
S3DIS & 0.76                 & 0.85                 & 0.96              & \textbf{0.99}               & \textbf{0.89}                 & 0.90                \\ 
ScanNet & 0.77                 & 0.85                 & 0.96                & \textbf{0.99}               & \textbf{0.89}                 & 0.90                \\ 
S3DIS+ScanNet & 0.79                 & 0.86                & 0.96            & \textbf{0.99}              & \textbf{0.89}                 & \textbf{0.91}               \\
S3DIS+ScanNet+3DSSG & \textbf{0.80}                 & \textbf{0.87}                & \textbf{0.97}            & \textbf{0.99}              & \textbf{0.89}                 & \textbf{0.91}                \\
           \bottomrule
\end{tabular}
\caption{\textbf{Pre-training dataset.} A comparison of \ours pre-trained on different datasets. The datasets are ordered by size.}
\label{tab:ablation_dataset}
\vspace{-0.5cm}
\end{table}
}

\newcommand{\tabfigpercenteval}{
\begin{table}[t]
\tabcolsep=4pt
\centering
\small
\begin{subtable}{\linewidth}
\includegraphics[width=0.97\linewidth]{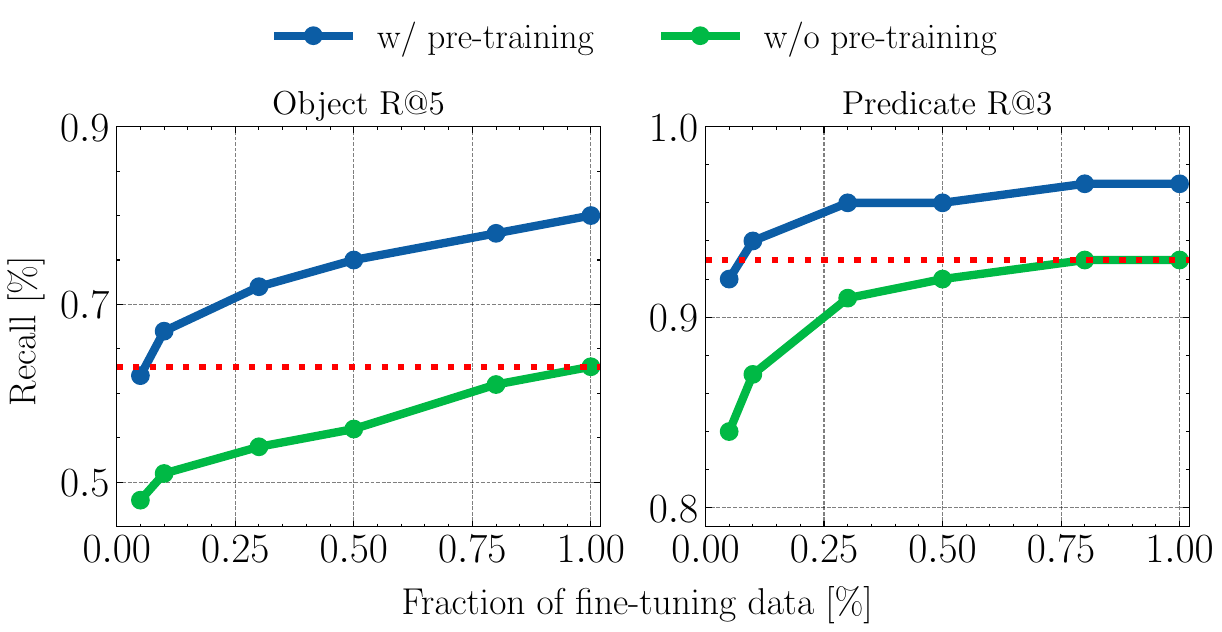}
\caption{Comparison of \ours with and without pre-training under limited labeled fine-tuning data.}
\label{tab:figablation_trainingdata}
\end{subtable}
\\
\
\\
\vspace{-0.2cm}
\begin{tabular}{@{}lcccccc@{}}
\toprule
       & \multicolumn{2}{c}{Object} & \multicolumn{2}{c}{Predicate} & \multicolumn{2}{c}{Relationship} \\
       \cmidrule(lr){2-3}\cmidrule(lr){4-5}\cmidrule(lr){6-7}
      Method & R@5                  & R@10                 & R@3                 & R@5                & R@50                 & R@100                \\ \midrule
$\text{\ours}_{100\%}$ & \textbf{0.80}                 & \textbf{0.87}                 & \textbf{0.97}                & \textbf{0.99}               & \textbf{0.89}                 & \textbf{0.91}                \\ 
$\text{\ours}_{80\%}$ & 0.78                 & 0.86                & 0.94            & 0.98               & 0.89                 & 0.90                \\
$\text{\ours}_{50\%}$ & 0.75                 & 0.83                 & 0.93                         & 0.97               &  0.88                & 0.89                \\
$\text{\ours}_{30\%}$ & 0.72                 & 0.82                 & 0.92                         & 0.97               &  0.87                & 0.88                \\
$\text{\ours}_{10\%}$ & 0.67                 & 0.75                 & 0.90                         & 0.95               &  0.84                & 0.85                \\
$\text{\ours}_{05\%}$ & 0.62                 & 0.71                 & 0.89                & 0.93               & 0.81                & 0.83                \\ 
\bottomrule
\end{tabular}
\begin{subtable}{\linewidth}
\caption{Affects of limited labeled fine-tuning data for \ours.}
\label{tab:tabablation_trainingdata}
\end{subtable}
\vspace{-0.5cm}
\caption{\textbf{Limited fine-tuning data.} A comparison of \ours fine-tuned using different quantities of 3DSSG labels.}
\label{tab:ablation_trainingdata}
\vspace{-0.4cm}
\end{table}
}

\newcommand{\newtablepretrainevaluation}{
\begin{table}[t]
\tabcolsep=0.75mm
\centering
\small
\begin{tabular}{@{}lccccccc@{}}
\toprule
       & &\multicolumn{2}{c}{pre-train}  & \multicolumn{2}{c}{Object} & \multicolumn{2}{c}{Predicate}\\
       \cmidrule(lr){3-4}\cmidrule(lr){5-6}\cmidrule(lr){7-8}
       & GCN & PCL & SG & R@5            & mR@5        & R@3                      & mR@3                    \\
      \midrule
      
      STRL \cite{Choy_2019_CVPR} &  & \checkmark &  & 0.75   & 0.35  & 0.94  & 0.50 \\
      STRL \cite{Choy_2019_CVPR} & \checkmark & \checkmark &  & 0.63   & 0.23  & 0.92  & 0.48 \\
      DepthContrast \cite{Zhang_2021_ICCV} & & \checkmark &  & 0.77  & 0.36  & 0.94   & 0.51 \\
      DepthContrast \cite{Zhang_2021_ICCV} & \checkmark & \checkmark &  & 0.60  & 0.22  & 0.93   & 0.50 \\
      \midrule
      Ours (no pre-train) &\checkmark &  &  & 0.63  & 0.30   & 0.94  & 0.57 \\
      Ours (no GCN) & &  & \checkmark & 0.75 & 0.31 & 0.94 & 0.48\\
      Ours   &\checkmark &  & \checkmark  & \textbf{0.80} & \textbf{0.45} & \textbf{0.97} & \textbf{0.69}\\
\bottomrule
\end{tabular}
\caption{\textbf{Pre-training comparison.} We compare \ours with existing recent point cloud-based pre-training approaches (PCL). Our novel scene graph pre-training (SG) shows high effectiveness outperforming point cloud-based pre-training approaches. Adding a GCN to our method is beneficial since we optimize it during pre-training, while point cloud-based pre-training approaches do not benefit from a GCN which is not pre-trained.}
\label{tab:pretrain_evaluation}
\vspace{-0.5cm}
\end{table}
}

\newcommand{\tabheadbodytail}{
\begin{table}[t]
\tabcolsep=2.8mm
\centering
\small
\begin{tabular}{@{}lccccc@{}}
\toprule
      & & Head & Body & Tail & All\\
       \midrule
       \multirow{2}{*}{Objects} &w/o pre-train & 0.88 & 0.45 & 0.06 & 0.30 \\
       & w/ pre-train & 0.92 & 0.78 & 0.24 & 0.45 \\
       \midrule
       \midrule
       \multirow{2}{*}{Predicates} & w/o pre-train & 0.94 & 0.83 & 0.41 & 0.57 \\
       & w/ pre-train & 0.97 & 0.96 & 0.65 & 0.69 \\
\bottomrule
\end{tabular}
\caption{\textbf{Frequency based class evaluation.} We sort objects and predicates into head, body, tail classes based on their occurrence and compare our pre-training method with the same model not pre-trained. We compare on the mR@5 metric objects and the mR@3 metric for predicates.}
\label{tab:headbodytail}
\vspace{-0.5cm}
\end{table}
}

%% file: preamble.tex
\definecolor{turquoise}{cmyk}{0.65,0,0.1,0.3}
\definecolor{purple}{rgb}{0.65,0,0.65}
\definecolor{dark_green}{rgb}{0, 0.5, 0}
\definecolor{orange}{rgb}{0.8, 0.6, 0.2}
\definecolor{red}{rgb}{0.8, 0.2, 0.2}
\definecolor{darkred}{rgb}{0.6, 0.1, 0.05}
\definecolor{blueish}{rgb}{0.0, 0.3, .6}
\definecolor{light_gray}{rgb}{0.7, 0.7, .7}
\definecolor{pink}{rgb}{1, 0, 1}
\definecolor{greyblue}{rgb}{0.25, 0.25, 1}

\definecolor{thirdbestcolor}{rgb}{1,1, 0.6}
\definecolor{secondbestcolor}{rgb}{1, 0.9, 0.6}
\definecolor{firstbestcolor}{rgb}{1, 0.6, 0.6}

\usepackage[inline]{enumitem}
\setlist[itemize]{noitemsep,leftmargin=.15in,topsep=.2em}
\setlist[enumerate]{noitemsep,leftmargin=*,topsep=.2em}

\newcommand{\Loss}[1]{\mathcal{L}_\text{#1}}
\newcommand{\Lbbox}{\Loss{bbox}}
\newcommand{\Langle}{\Loss{angle}}
\newcommand{\Lshape}{\Loss{shape}}

\newcommand{\Lsg}{\Loss{SG}}

\newcommand{\enumlabel}{\textbf{(\roman*)}}

\usepackage[acronym]{glossaries}
\usepackage{glossaries-extra}
\newacronym{spo}{s-p-o}{subject-predicate-object}
\newacronym{gcn}{GCN}{Graph Convolutional Network}
\newacronym{mlp}{MLP}{Multi Layer Perceptron}

\newcommand{\new}[1]{#1}
\newcommand{\critical}[1]{#1}

\usepackage{multirow}
\usepackage[accsupp]{axessibility}

%% file: title.tex
\newcommand{\ours}{SGRec3D\xspace} %
\newcommand{\thetitle}{\ours: Self-Supervised 3D Scene Graph Learning \\via Object-Level Scene Reconstruction}
\newcommand{\thetitleplain}{\ours: Self-Supervised 3D Scene Graph Learning via Object-Level Scene Reconstruction} %

%% file: content/abstract.tex
\begin{abstract}
In the field of 3D scene understanding, 3D scene graphs have emerged as a new scene representation that combines geometric and semantic information about objects and their relationships.
However, learning semantic 3D scene graphs in a fully supervised manner is inherently difficult as it requires not only object-level annotations but also relationship labels. 
While pre-training approaches have helped to boost the performance of many methods in various fields, 
pre-training for 3D scene graph prediction has received little attention. 
Furthermore, we find in this paper that classical contrastive point cloud-based pre-training approaches are ineffective for 3D scene graph learning.
To this end, we present \ours, a novel self-supervised pre-training method for 3D scene graph prediction. We propose to reconstruct the 3D input scene from a graph bottleneck as a pretext task.
Pre-training \ours does not require object relationship labels, making it possible to exploit large-scale 3D scene understanding datasets, which were off-limits for 3D scene graph learning before.
Our experiments demonstrate that in contrast to recent point cloud-based pre-training approaches, our proposed pre-training improves the 3D scene graph prediction considerably, which results in SOTA performance, outperforming other 3D scene graph models by \textbf{+10\%} on object prediction and \textbf{+4\%} on relationship prediction.
Additionally, we show that only using a small subset of 10\% labeled data during fine-tuning is sufficient to outperform the same model without pre-training.\looseness=-1

\end{abstract}

%% file: content/introduction.tex
\figteaser
\vspace{-1.5em}
\section{Introduction}
\label{sec:intro}
Scene graphs provide a graph-based representation of a scene, by not only representing the geometric scene objects, but also their relation among each other. In recent years, 2D scene graphs have seen a wide range of applications in computer vision  and robotics~\cite{Johnson_2018_CVPR,bielecki2016graph,Ashual_2019_ICCV,Dhamo_2020_CVPR}. Consequently, many approaches for generating 2D scene graphs based on given input images have been proposed~\cite{woo2018linknet,yin2018zoom,suhail2021energy}. 
In the same way as 2D scene graphs capture structured knowledge about scenes represented through images, 3D scene graphs can capture the same information for point clouds or other 3D data structures.
Despite the fact that 3D scene graphs are widely used in computer graphics~\cite{granskog2021neural}, and despite their great potential for solving computer vision or robotics tasks~\cite{Rosinol_2020_RSS,Wu_2021_CVPR,hughes2022hydra,Agia_2022_PMLR}, relatively little work has been done to predict 3D scene graphs based on a given 3D scene.

Predicting 3D scene graphs comes with several challenges on its own. It does not only have to provide a high level representation of a given 3D scene, but it must also derive this from often noisy and incomplete sensor data.
Thus, 3D scene graph generation is difficult to tackle with rule-based deterministic algorithms. For instance, two chairs of the \textit{same style} could have different visual appearances or a jacket \textit{lying on} one chair could occlude most of the visible surface.
While first approaches have been proposed to learn a 3D scene graph based on a 3D point cloud~\cite{Wald_2020_CVPR,Zhang_2021_CVPR,Zhang_2021_Neurips,wald2022learning}, these approaches require labels to be available, as they perform scene graph prediction in a fully supervised manner. 
Acquiring data and labels in 3D is very challenging task and requires extensive human effort, which is underlined by the particular scarcity of labeled training data in the domain of 3D scene graphs. 
Therefore, our goal is to reduce the need for such labels when learning to predict 3D scene graphs.

In recent years, self-supervised pre-training methods in 3D have shown to be effective in improving results of neural architectures with a high data demand by utilizing the available data more efficiently without requiring additional annotated data \cite{Hou_2021_CVPR,Chen_2022_ECCV,Xie_2020_ECCV,He_2022_CVPR}.
\critical{
Despite the promising nature of pre-training approaches in low data regimes, self-supervised pre-training for 3D scene graphs has not been investigated so far. In particular, we find in this paper that point cloud-based pre-training approaches are ineffective for 3D scene graphs (see \cref{sec:experiments}).}
\new{
To solve this issue, we propose a new self-supervised pre-training approach tailored for 3D scene graphs using an encoder-decoder-based method with a graph bottleneck and a graph-based pretext task. We choose graph-based 3D reconstruction as our pretext task which, unlike previous point cloud-based pre-training approaches, considers the graph information directly to learn the optimal information flow through the graph to reconstruct 3D scene point clouds.
In contrast to 2D, 3D scene reconstruction remains a challenging problem, mainly due to the sparsity and non-continuous nature of 3D point clouds.
}

Thus, our main contributions are: 
\new{
\begin{enumerate*}[label=\enumlabel]
    \item We propose a novel self-supervised pre-training method designed for 3D scene graph predictions. To the best of our knowledge, this is the first pre-training approach designed for 3D scene graphs.
    \item We demonstrate how to utilize additional 3D datasets to boost the effectiveness of our pre-training approach without being dependent on scene graph labels.
    \item We outperform fully-supervised methods and \critical{our novel pre-training shows greater effectiveness than other point cloud-based pre-training baselines.}
    \item Our pre-trained method demonstrates significantly improved label efficiency by requiring only 5\%-10\% of scene graph labels to outperform the same model trained from scratch on a complete labeled dataset.
\end{enumerate*}
}

%% file: content/related_work.tex
\figmainmethod
\vspace{-0.5cm}
\section{Related Work}
\label{sec:rel_work}

\noindent\textbf{Scene graph prediction.} A scene graph is a data structure that represents a scene as a graph, where nodes provide semantic descriptions of objects in the scene and edges represent relationships between objects. In computer vision, scene graphs were first introduced by Johnson~\etal~\cite{Johnson_2015_CVPR} motivated by image retrieval. Subsequent works focused primarily on the refinement of scene graph prediction from images~\cite{Li_2017_ICCV,Herzig_2018_NEURIPS,Yang_2018_ECCV,Zellers_2018_CVPR,Li_2018_ECCV}, while utilizing different methods such as message passing~\cite{Xu_2017_CVPR}, \gls{gcn}~\cite{Kipf_2017_ICLR} or attention~\cite{Qi_2019_CVPR}. Some works also investigate the incorporation of prior knowledge into the graph learning problem~\cite{Chen_2019_CVPR,Sharifzadeh_2021_AAAI}. Much of the progress is accounted to the introduction of visual gnome~\cite{krishna_2017_IJCV}, a large scale dataset for connecting language and vision which contains scene graph annotations for images. Chang~\etal~\cite{chang2021scene} provide a comprehensive survey of scene graph generation approaches and their applications.
Other works, instead, apply semantic scene graphs to image generation~\cite{Johnson_2018_CVPR,Ashual_2019_ICCV}, and image manipulation~\cite{Dhamo_2020_CVPR}.  

Applications of scene graphs can be also found in the 3D domain where literature presents two main approaches which explore their potential. Wald~\etal~\cite{Wald_2020_CVPR} introduce the first 3D scene graph dataset 3DSSG, with focus on semantics with 3D graph annotations, build upon the 3RScan dataset~\cite{Wald_2019_ICCV}. Based on this dataset, subsequent works extended the common principles of 2D scene graph prediction to 3D~\cite{Wald_2020_CVPR,wald2022learning}. 
Other works explore unique approaches for 3D scene graphs utilizing novel graph neural networks \cite{Zhang_2021_CVPR}, transformers \cite{lv2023revisiting}, the use of prior knowledge~\cite{Zhang_2021_Neurips}, or image-based oracle models \cite{wang2023vl}. Others explore applications utilizing 3D scene graphs for 3D scene generation and manipulation \cite{Dhamo_2021_ICCV}, the alignment of 3D scans with the help of 3D scene graphs \cite{sarkar2023sgaligner}, or dynamic construction of 3D scene graphs~\cite{Wu_2021_CVPR,Wu_2023_CVPR} during the exploration of a 3D scene.
In contrast, our approach focuses on a novel pre-training strategy for scene graph prediction, without requiring additional scene graph labels.

\noindent\textbf{Pre-training for 3D scene understanding.} Deep learning methods are well known for requiring large amounts of training data. Since collecting data and providing labels is costly and time-consuming, pre-training methods have emerged in the field of scene understanding. In the 2D domain, for instance, 
pre-training on existing large-scale datasets, such as ImageNet~\cite{Deng_2009_CVPR}, is a common practice.
More recently, methods such as masked autoencoders~\cite{He_2022_CVPR} have demonstrated, that pre-training alone on the target dataset using a pretext task can improve results by a considerable margin. In 3D, representation learning approaches demonstrate that using only a fraction of available point labels can lead to similar results as obtainable with fully supervised methods when pre-trained with a self-supervised pretext task~\cite{Hou_2021_CVPR,Zhang_2021_ICCV,Xie_2020_ECCV, Huang_2021_ICCV,Chen_2022_ECCV}.
However, so far neither of these works have considered 3D scene graph prediction as the downstream task. In this work, we will compare existing pre-training methods designed for point cloud pre-training with our approach designed for 3D scene graph learning.\looseness=-1

\noindent\textbf{3D scene reconstruction.} Literature shows a number of methods able to generate 3D scenes from images~\cite{Nie_2020_CVPR,Zhang_2021_CVPR,Gkioxari_2022_CVPR}. Other works aim to complete a 3D scene from an incomplete 3D scan~\cite{Dai_2018_CVPR,Dai_2020_CVPR,Siming_2022_CORR}. But only few works attempt to do full 3D scene reconstruction from point clouds~\cite{Peng_2020_ECCV}, and most methods for 3D reconstruction are limited to object reconstructions~\cite{Groueix_2018_CVPR,Yang_2019_ICCV,Cai_2020_ECCV,Luo_2021_CVPR} on datasets like ShapeNet~\cite{Chang_2015_CORR} or ModelNet~\cite{Wu_2015_CVPR}. Methods more similar to our approach explore 3D scene generation from graphs~\cite{Wang_2019_ACM,Luo_2021_CVPR}, however most methods simplify the task of 3D generation. Li~\etal~\cite{Li_2019_ACM} for instance introduce GRAINS, a recursive VAE to generate a 3D layout followed by object retrieval to synthesize a 3D indoor scene. Dhamo~\etal~\cite{Dhamo_2021_ICCV} go beyond object retrieval and attempt to generate and manipulate 3D scenes by reconstructing objects individually from a scene graph using a generative graph-based model.
Similar to this work, we design our decoder to reconstruct the 3D scene from a graph bottleneck. However, in contrast to this work, we reconstruct the input scene, instead of generating plausible object shapes and layouts.

%% file: content/method.tex
\section{Method}
\label{sec:method}
We propose \ours, a novel pre-training method to learn 3D scene graphs from 3D data in an autoencoder-like manner, as shown in \cref{fig:short}. Like all autoencoding approaches, our method consists of an encoder that maps the input to a latent representation and a decoder that reconstructs the original input from the latent representation. 
But unlike most autoencoder approaches, our method maintains a graph representation within the network given a non-graph input and output.
The encoder (see \cref{sec:method:encoder}), takes as input a point cloud partitioned using object instances and their bounding boxes.
From this input, the encoder generates a minimal representation as a 3D scene graph in a graph bottleneck (see \cref{sec:method:bottleneck}), by learning to reconstruct from this representation the input scene using a decoder (see \cref{sec:method:decoder}). This pre-trained architecture can then be fine-tuned to predict a semantic 3D scene graph $\mathcal{G}=\{\mathcal{N},\mathcal{E}\}$, where nodes $\mathcal{N}$ represent object instances within a corresponding 3D point cloud, while edges $\mathcal{E}$ express predicates that form together with object nodes semantic relationships (see \cref{sec:method:training}). Each edge in the graph can represent zero or more relationships.

\subsection{Encoder}
\label{sec:method:encoder}
\new{
Given a point cloud $\mathcal{P}$ of a scene $\mathcal{S}$ with class-agnostic instance segmentation $\mathcal{M}$ provided by an off-the-shelf instance segmentation method such as Mask3D~\cite{Schult_2022_CORR} or a dataset, we extract each point set $\mathcal{P}_i$ containing instance $i$ and its axis-aligned or oriented bounding box $\mathcal{B}_i$ using the mask $\mathcal{M}_i$. Moreover, for every instance pair \mbox{$\langle i, j \rangle \in \lVert \mathcal{M} \rVert \times \lVert \mathcal{M} \rVert$}, we get the point set $\mathcal{P}_{ij}$ using the union of their respective bounding boxes $\mathcal{B}_{ij} = \mathcal{B}_{i} \cup \mathcal{B}_{j}$. Note that the point set $\mathcal{P}_{ij}$ contains not only the union of the masked instances $\mathcal{P}_{i} \cup \mathcal{P}_{j}$, but also other points falling within the volume $\mathcal{B}_{ij}$. This helps to augment the point cloud with contextual information relating the two objects. $\mathcal{P}_{i}$, $\mathcal{B}_{i}$ and $\mathcal{P}_{ij}$ serve as input to our scene encoder.
}
The encoder follows the common principles of scene graph prediction from prior 2D and 3D works~\cite{Lu_2016_ECCV,Xu_2017_CVPR,Yang_2018_ECCV,Wald_2020_CVPR}. 
We construct an initial graph with node features $\phi_n$ and edge features $\phi_p$ from the extracted instance and bounding box features. Each point set $\mathcal{P}_i$ is fed into a shared PointNet~\cite{Qi_2017_CVPR} to extract features for object nodes. Every point set $\mathcal{P}_{ij}$ is concatenated with a mask which is equal to $1$ if the point corresponds to object $i$, $2$ if the object corresponds to object $j$, and $0$ otherwise. The concatenated feature vector is then fed into another shared PointNet to extract features for predicate edges. Additionally, the centers of the point sets $\mathcal{P}_i$ and $\mathcal{P}_{ij}$ are normalized before inputting them into the respective PointNet.

The extracted node and edge features are then arranged as triplets \mbox{$t_{ij} = \langle \phi_{n,i},\phi_{p,ij},\phi_{n,j} \rangle$} in a graph structure. This initial feature graph is passed into a \gls{gcn}~\cite{Kipf_2017_ICLR}. Each \gls{gcn} layer $l_g$ processes the triplets $t_{ij}$ and propagates the information through the graph in three steps, with a similar message passing procedure to \cite{Wald_2020_CVPR}.
First, $t_{ij}$ is fed into a MLP $g_1(\cdot)$
\begin{equation}
    (\psi_{n,i}^{(l_g)},\phi_{p,ij}^{(l_g+1)},\psi_{n,j}^{(l_g)}) = g_1\left(\phi_{n,i}^{(l_g)},\phi_{p,ij}^{(l_g)},\phi_{n,j}^{(l_g)}\right)
    \label{eq:gcn_propagate}
\end{equation}
where $\psi$ represents the nodes' processed features. After this first pass, the resulting edge feature $\phi_{p,ij}^{(l_g+1)}$ does not need any further refinement.\\
Second, an aggregation function averages the incoming information from all the connected edges of each node
\begin{equation}
    \rho_{n,i}^{(l_g)} = \frac{1}{N_i}\left(\sum_{k\in \mathcal{R}_i}\psi_{n,k}^{(l_g)} + \sum_{k\in \mathcal{R}_j}\psi_{n,k}^{(l_g)}\right)
    \label{eq:gcn_aggregate}
\end{equation}
where $N_i$ denotes the number of edges connected to node $i$, and $\mathcal{R}_i$ and $\mathcal{R}_j$ are the set of nodes connected to node $i$ and node $j$ respectively.\\
Finally, the resulting node feature $\rho_{n,i}^{(l_g)}$ passes into a second update MLP $g_2(\cdot)$ and a residual connection is added:
\begin{equation}
    \phi_{n,i}^{(l_g+1)} = \phi_{n,i}^{(l_g)} + g_2(\rho_{n,i}^{(l_g)}).
    \label{eq:gcn_residual}
\end{equation}
In the end, the processed features \(\phi_{n,i}^{(l_g+1)},\phi_{p,ij}^{(l_g+1)},\phi_{n,j}^{(l_g+1)}\) are passed to the next layer of the network.

\subsection{Graph bottleneck}
\label{sec:method:bottleneck}
Features are further processed through multiple layers of graph convolutions, propagating them to neighboring nodes.
A final MLP $f_n(\cdot)$ is applied to all node features, and a $softmax$ activation function represents the nodes as a probability distribution over the node classes
\begin{equation}
    \phi_{n,i}^{(e)} = softmax(f_n(\phi_{n,i}^{(l_n)})).
    \label{eq:node_embedding}
\end{equation}
where $\phi_{n,i}^{(e)}$ is the final encoder feature vector for each node which is passed into the decoder. 

The edge features, instead, are handled by a different MLP $f_p(\cdot)$ and by a class-wise sigmoid activation function to map the edges to a separate probability distribution for each possible relationship between object nodes
\begin{equation}
    \phi_{p,i}^{(e)} = \sigma(f_p(\phi_{p,i}^{(l_n)})).
    \label{eq:edge_embedding}
\end{equation}
where $ \phi_{p,i}^{(e)}$ is the final encoder feature vector for each edge which is passed into the decoder.

\subsection{Decoder}
\label{sec:method:decoder}
The goal of our scene decoder is to reconstruct the original scene from the bottleneck scene graph representation. 
To preserve the layout and object details, we first pass the low-dimensional features into an embedding MLP which lifts the latent graph representation to a high-dimensional feature-space.
Then, we further decode the latent graph using another \gls{gcn} with the same message passing structure as the encoder.  
Due to the low-dimensionality of the bottleneck and the ambiguity of the scene graph, the decoding step may be affected by information loss. 
Thus, we address this problem by introducing an additional skip-connection between the last \gls{gcn} encoder layer before applying the softmax and sigmoid functions and the first \gls{gcn} decoder layer by concatenating ($\oplus$) the GCN features with the embedded feature from the bottleneck. 
For the node and edge features this is defined as follows
\begin{equation}
\small
    \phi_{n,i}^{(d_{in})} = (\phi_{n,i}^{(e)} \oplus \phi_{n,i}^{(l_n)}), \quad \phi_{p,i}^{(d_{in})} = (\phi_{p,i}^{(e)} \oplus \phi_{p,i}^{(l_n)}).
    \label{eq:decoder_input}
\end{equation}
where $\phi_{n,i}^{(d_{in})}/\phi_{p,i}^{(d_{in})}$ are the input decoder features for each node and edge.

Reconstructing a full 3D scene is a highly complex task, giving the sparsity of 3D data. 
Therefore, we choose to reconstruct each object individually rather than the complete scene. To this end, we combine the 3D bounding box of each object, predicted by the Box-Head, with the corresponding object reconstruction provided by the Shape-Head.
For the final scene reconstruction, we place each generated object within its matching bounding box.\looseness=-1

For the Box-Head, we implement an MLP to predict the box extents $[h,w,d]$ and the center location $[c_x,c_y,c_z]$. Predicting the 3D orientation of objects using regression has shown to be difficult given the non-linearity of the 3D rotation space \cite{Mahendran_2017_ICCV}.
Therefore, we predict the object's orientation angle $\alpha$ separately by means of classifying it into 1 out of 24 discrete bins, rather than regressing the angle directly. 
The Shape-Head consists of an MLP that predicts a 1D latent vector which is further processed by the decoder of AtlasNet~\cite{Groueix_2018_CVPR} pre-trained on ShapeNet~\cite{Chang_2015_CORR}, which reconstructs the object from the latent vector. 

\subsection{Pre-training using scene reconstruction}
\label{sec:method:training}
For pre-training we learn to reconstruct the 3D scene by predicting the bounding box and the shape of the objects. The loss for the object-level scene reconstruction is composed of three components: 
\begin{enumerate*}[label=\enumlabel]
    \item a bounding box regression loss $\Lbbox$ which uses the $L_1$ distance for the bounding box parameters,
    \item a cross entropy classification loss $\Langle$, and
    \item an $L_1$ loss $\Lshape$ for the shape embedding before applying the AtlasNet decoder:
\end{enumerate*}
\begin{equation}
    \mathcal{L}_\text{rec} = \eta_1 \mathcal{L}_\text{bbox} + \eta_2 \mathcal{L}_\text{angle} + \eta_3 
    \mathcal{L}_\text{shape}
    \label{eq:rec_loss_simplified}
\end{equation}
where $\eta_i$ are weighting factors.
Note that this loss does not rely on scene graph labels which allows for the use of additional training data from larger 3D data sets, 
as we will demonstrate in Section~\ref{sec:experiments}.

After pre-training, our model needs to be fine-tuned on the downstream task of predicting 3D scene graphs. For this, we discard the decoder and fine-tune the pre-trained encoder using the scene graph annotations 
with a fully supervised loss $\Lsg$. 
It consists of a \mbox{cross-entropy} loss $\mathcal{L}_{obj}$ for the node classification and a per-class \mbox{binary cross entropy} loss $\mathcal{L}_{pred}$ for the predicate prediction. The latter is used to learn different predicates separately from one another to support multi-predicate relationships. The combined loss is defined as
\begin{equation}
    \mathcal{L}_\text{SG} = \lambda_1 \mathcal{L}_\text{obj} + \lambda_2 \mathcal{L}_\text{pred}
    \label{eq:sg_loss_simplified}
\end{equation}
where $\lambda_1$ and $\lambda_2$ are the respective weighting factors.

Further details and documentation of our model architecture, training procedure, chosen loss functions and weighting factors are provided in the supplementary material.

%% file: content/experiments_new.tex
\section{Experiments}
\label{sec:experiments}

\figexamples

\subsection{Experimental setup}
\noindent\textbf{Datasets.} To prove the effectiveness of our proposed method, we evaluate it on real-world 3D scans from the 3DSSG \cite{Wald_2020_CVPR} dataset.
3DSSG is currently the only real-world dataset that provides semantic 3D scene graph annotations. Another 3D scene graph dataset is \cite{Armeni_2019_ICCV}, however, the scene graphs modeled in this dataset focus on hierarchical structuring and lack semantic relationship labels. In contrast, 3DSSG provides 3D scene graph labels for 160 distinct object classes and 27 relationship categories, corresponding to over 1,000 3D indoor point cloud reconstructions. 
The 3D scene graphs present in 3DSSG are further split into smaller sub-graphs spanning a small selection of objects per scene, yielding over 4,000 samples for training and evaluation. We follow the previous work~\cite{Wald_2020_CVPR} and use the same scene graph and training/evaluation splits first introduced by Wald \etal~\cite{Wald_2019_ICCV}.
The 3DSSG dataset, however, is a rather small dataset, including only 478 different scenes, which may be challenging for training large deep learning architectures. To alleviate this problem, we additionally pre-train on existing indoor object detection datasets ScanNet \cite{Dai_2017_CVPR} and S3DIS \cite{Armeni_2017_CORR}. ScanNet and S3DIS are much larger indoor datasets, including 1513 and 727 annotated scenes respectively. The available baselines cannot use additional datasets since they require ground truth scene graph annotations. In contrast, our pretext task does not require these annotations, which makes it possible for us to utilize additional datasets for pre-training.

\noindent\textbf{Evaluation metrics.} Following previous works \cite{wald2022learning,Zhang_2021_CVPR,Wald_2020_CVPR,Xu_2017_CVPR,Yang_2018_ECCV}, we evaluate object node classification and predicate edge prediction separately. To analyze the overall scene graph prediction performance, we jointly compute the accuracy of relationships consisting of triples formed by two nodes (subject \& object) and their connecting edge (predicate). Since we predict object nodes and predicate edges independently, we adapt the approach first introduced by Yang \etal~\cite{Yang_2018_ECCV} for relationship evaluation. Through this method, we obtain a scored list of triplet predictions by multiplying the object node confidences with the predicate edge probability. For comparison with previous works, we follow \cite{wald2022learning,Zhang_2021_CVPR,Wald_2020_CVPR} and use the top-k recall metric first introduced by Lu \etal~\cite{Lu_2016_ECCV} for scene graph prediction.
\new{
For objects and predicates, we further do a class-wise evaluation by splitting the classes based on the frequency of number of labels in the train set into head, body and tail respectively. A class-wise evaluation over all categories as well as for the head, body, and tail splits enables a more precise understanding of the prediction performance. For this we use the same top-k metric formulation, which is also known as the more precise but less commonly used mR@k metric.
}

\tabmaineval
\subsection{3D scene graph prediction}
\noindent \textbf{Comparison with fully supervised methods.}
To show the impact of our pre-training, we compare it against recent fully supervised 3D scene graph baselines (SGGPoint~\cite{Zhang_2021_CVPR}, 3DSSG~\cite{Wald_2020_CVPR}, SGFN~\cite{Wu_2021_CVPR} and \mbox{Liu \etal~\cite{Liu_2022_TVCG}}) and adopted 2D scene graph methods (MSDN~\cite{Li_2017_ICCV}, KERN~\cite{Chen_2019_CVPR} and BGNN~\cite{Li_2021_CVPR}). For the 2D scene graph methods, the 2D object detector was replaced by a PointNet-based feature extractor. We note that, to alleviate the severe object class imbalance in the scene graph prediction task, SGGPoint only provides a model for 27 object classes and 16 relation classes in the 3DSSG dataset.

Results in \cref{tab:main_evaluation} show that \ours outperforms most existing fully supervised methods by a large margin, and our closest competitor SGFN~\cite{Wu_2021_CVPR} by a considerable amount. 
Especially on object node classification \ours outperforms SGFN by a large margin (+10\%/+7\%). 
For relationship prediction, we also report favorable results with a +4\% increase to SGFN on both metrics.
For the predicate edge prediction, we observe similar results as SGFN. Given the overall high score for this metric, we assume that we reached a saturation point for this task on this dataset. In \cref{fig:scene_graph_visualisation}, we provide predicted 3D scene graphs for three different scenes.
Our method is able to predict accurate and mostly correct scene graphs for the given scenes. Objects are predicted well, with most nodes being predicted correctly and only some nodes being predicted incorrectly, where our method often chooses an object class of a similar meaning. Similarly, predicates between objects are also predicted with a high accuracy with only a few false positive predictions.

\tabheadbodytail
\noindent \textbf{Class-wise evaluation.}
\new{
To further investigate the impact of our proposed pre-training, in \cref{tab:headbodytail} we provide a detailed comparison of our method with and without pre-training for individual classes grouped into head, body and tail based on their frequency. Additionally, we provide \textit{All} which is the average recall over all classes individually also known as the mR@k metric. The improvement of our pre-training over all classes is large with a +27\% gain for object classification and a +24\% gain for predicate prediction on the mR@k metric. 
We observe, that this improvement originates mostly from a very large improvement on rare body and tail classes. The improvement on more frequent head classes is smaller since the baseline method already produces good results for frequent categories.
}
\newtablepretrainevaluation
\figreconstructions

\noindent \textbf{Comparison with point cloud-based pre-training.}
We are the first to investigate pre-training designed for 3D scene graphs by considering the graph nature of scene graphs during pre-training.
In \cref{tab:pretrain_evaluation}, we show a comparison with recent point cloud-based pre-training approaches. In contrast to our method, these approaches do not model the graph structure of the scene graph during pre-training.
We compare with the pre-trained 3D feature encoders from both STRL~\cite{Choy_2019_CVPR} and DepthContrast \cite{Zhang_2021_ICCV}. 
We choose these two approaches because they have been proven to be highly effective in pre-training 3D scene understanding models for tasks such as 3D segmentation and detection. Similar to our method, they rely on a PointNet++ feature extraction backbone and ScanNet as pre-training data.
We add two prediction heads on top of their pre-trained PointNet++ backbones for objects and predicates and fine-tune them on the 3DSSG dataset. 
For further comparisons, we add a GCN between the pre-trained backbone and the prediction heads to make their network architecture very similar to ours. 
The major difference is that while our GCN contains pre-trained weights, their GCN is randomly initialized because only the 3D feature extractors can be pre-trained by STRL and DepthContrast.
\cref{tab:pretrain_evaluation} demonstrates, that our scene graph pre-training (SG) tailored for 3D scene graph prediction produces drastically better results than our point cloud-based pre-training baselines (PCL), with 9\% improvement on the mR@5 metric for objects and 18\% on the mR@3 metric for predicates. We observe that our novelty of a pre-trained graph neural network greatly improves the pre-training effectiveness of our method
(+14\%~object, +11\% predicate prediction), the same is not true for the point cloud-based pre-training methods. 
We assume this is because, in contrast to our method where the graph layers are optimized during pre-training, the point cloud-based approaches do not optimize the graph. Adding the graph layers during fine-tuning consequently adds a considerable number of untrained weights which are challenging to train on a rather small dataset such as 3DSSG.

\vspace{-0.5cm}
\subsubsection{Scene reconstruction}
The benefit of our pre-training is influenced by the ability of our model to learn the reconstruction pretext task. Since our downstream task includes learning relationships in scene graphs, it is crucial that the relationships present in the original scene remain preserved in the reconstructed scene. This indicates that the model learns transferable knowledge for the downstream scene graph prediction during pre-training.

\tabreconeval

\cref{fig:scene_reconstructions} shows some qualitative results for reconstructing 3D scene splits from 3DSSG. 
In general, our model correctly reconstructs the layouts of the scenes.
In all predicted scenes, the reconstructed objects are located in similar positions to the ones in the original scene.
Relationships that describe the relative proximity of objects are clearly preserved, such as \textit{Close by, Left, Right, \etc} Sometimes objects hanging on walls are generated on the wrong side of the wall (see \cref{fig:scene_graph_visualisation}a), however relationships like \textit{hanging on, attached to} are still maintained in the generation. The shape of the objects differs in detail compared to the original scenes. This is because we do not fine-tune the AtlasNet \cite{Groueix_2018_CVPR} decoder for more stable training. Still, the rough shape of the objects is preserved and relationships like \textit{same as, bigger than, smaller than} are maintained. 

\new{
In \cref{tab:rule_verify}, we provide a quantitative evaluation of the preserved relationships for those predicates where a simple rule can be approximated. We compare our results with the results from Graph-to-3D \cite{Dhamo_2021_ICCV}. 
The reported results confirm that our method preserves the original and reconstructed relationships in the scene. With an overall accuracy of 94\% the pretext task has been learned well by the model indicating an effective pre-training task. 
As shown in the table, we outperform Graph-to-3D, with great improvements for \textit{front of/behind of} relationships. 
While Graph-to-3D generates a scene solely using a scene graph as input, we reconstruct the scene from 3D using the graph bottleneck.
This retains more context information of the original scene compared to Graph-to-3D which tries to generate a novel scene. \looseness=-1
}
\vspace{-0.3em}
\subsection{Ablations}
\vspace{-0.2em}
\noindent\textbf{Pre-training dataset.}
Our method allows to leverage large-scale 3D datasets without scene graph labels during pre-training. 
In \cref{tab:ablation_dataset} we investigate the role of a larger pre-training dataset by reporting the fine-tuned performance of our method, given different pre-training datasets. 
We observe that pre-training on the 3DSSG \cite{Wald_2020_CVPR} dataset only, which is also used for fine-tuning, leads to competitive results compared to existing methods from \cref{tab:main_evaluation}. 
Pre-training on larger datasets like ScanNet \cite{Dai_2017_CVPR} and S3DIS \cite{Armeni_2017_CORR} further improves fine-tuning results. 
Finally, increasing pre-training data by combining individual datasets gives the best fine-tuned model. 
We can highlight considerable improvements in the object classification scores. The predicate scores improve only slightly, which we assume is correlated to a saturated metric. The relationship scores improve marginally with increasing dataset size, achieving best results by combining 3DSSG, ScanNet and S3DIS.

\tabdataseteval

\tabfigpercenteval

\noindent\textbf{Limited fine-tuning data.}
The goal of our proposed pre-training is to reduce the need for labeled scene graph data, which is often hard to annotate. 
To prove this contribution, in \cref{tab:tabablation_trainingdata} we provide an ablation of our pre-trained model fine-tuned on a fraction of labeled data from 3DSSG \cite{Wald_2020_CVPR}. 
We observe that reducing the number of labeled samples during fine-tuning only affects marginally the performance on object, predicate, and relationship prediction.
For example, using only a small fraction of the labeled data (\mbox{\raisebox{-0.8ex}{\~{}}10\%-30\%}) results in competitive performance compared to the works from \cref{tab:main_evaluation}. 
Moreover, we would like to emphasize that even with the 5\% of labeled data -- around 200 training samples -- \ours is able to achieve acceptable results.
In \cref{tab:figablation_trainingdata}, we compare the effects of limited labeled training data for our pre-trained model against the same model trained from scratch without pre-training. 
The pre-trained model outperforms the model trained from scratch on all data quantities by a large margin for object classification and predicate prediction. Furthermore, the pre-trained model requires only 5\%-10\% of labeled data to outperform the model trained from scratch.

%% file: content/conclusion.tex
\vspace{-0.15cm}
\section{Conclusion}
\label{sec:discussion}
\vspace{-0.1cm}
Pre-training for scene graphs has received little attention so far, despite its success for a variety of other downstream tasks. In this paper we find that existing point cloud-based pre-training approaches are ineffective for 3D scene graph prediction.
To this end, we present \ours, a novel self-supervised pre-training method for the downstream task of 3D scene graph prediction.
To the best of our knowledge, this is the first approach addressing pre-training for 3D scene graph prediction. 
Our experiments show that \ours significantly improves the predictions of objects and relationships in 3D scene graphs compared to existing fully supervised methods. We achieve these results thanks to
our pre-training contribution and the use of additional 3D datasets for pre-training.%
We show that even using a small percentage of limited fine-tuning data, \ours produces competitive results with recent methods.

%% file: content/supp.tex
This document supplements our work \textit{\thetitleplain} by providing 
\begin{enumerate*}[label=\enumlabel]
    \item reproducibility information on our implementation and training details (\cref{sec:impl}),
    \item more details on the dataset and its pre-processing (\cref{sec:dataset_details}),
    \item further ablations on our architecture design (\cref{sec:skip_connection}),
    \item a direct qualitative comparison between our method with and without pre-training for 3D scene graph prediction (\cref{sec:direct_comp}),
    \item additional 3D scene graph predictions using our method(\cref{sec:sg_resuls}),
    \item additional scene generations from our method (\cref{sec:scene_results}),
\end{enumerate*}

\section{Reproducibility Details}
\label{sec:impl}
\subsection{Network} 
Our encoder consists of two PointNets which pass features of size 256 to a 4-layer GCN, where $g_1(\cdot)$ and $g_2(\cdot)$ are composed of a linear layer followed by a ReLU activation.
Additionally, the bounding boxes of the object instances are encoded via a linear layer and are appended to the initial features from the PointNets.
The encoder GCN is followed by object and predicate prediction MLPs, consisting of 3 linear layers with batch normalization and ReLU activation. 

During pre-training, the resulting features are fed into the decoder part of our network, which consists of a 3-layer GCN with the same $g_1(\cdot)$, $g_2(\cdot)$ MLPs.
After the graph convolution, the GCN features are passed to
two different heads.
One is a Box-Head consisting of a 3-layer MLP with batch normalization and ReLU activation, outputting 7 box parameters ($w,l,h,c_x,c_y,c_z,\theta$). 
The other is a Shape-MLP with 3 linear layers, batch normalization and ReLU activation, outputting a 1024-dimensional shape latent code. The original object point clouds are additionally fed into the encoder of a pre-trained AtlasNet, which produces the target shape code for our model.

\subsection{Training} 
The model is trained with a batch size of 4, using an Adam optimizer with a learning rate of $0.0001$ and a reduce-on-plateau learning rate scheduler. The pre-training is performed until the validation loss converges. We pre-train our method for approximately 35 epochs until the validation loss for the reconstruction task converges.
Similarly, during fine-tuning, we monitor the validation loss. Once the validation loss converges, which occurs after \mbox{\raisebox{-0.8ex}{\~{}}20} epochs, we evaluate the scene graph prediction performance by calculating the metrics introduced in the paper on the validation set. Further training results in overfitting, indicated by the validation loss for both object prediction and predicate prediction. Overfitting occurs faster for the non-pre-trained model after around \mbox{\raisebox{-0.8ex}{\~{}}15} epochs. However, validation loss and evaluation metrics are worse than our pre-trained model. Further training with smaller learning rates does not improve the results.

\figcomparison

The training is performed on 2 NVIDIA Tesla V100 GPUs with 32 GB Memory. 

\subsection{Losses}
During pre-training we use the following reconstruction loss for all objects $i\in N$ in the scene 
\begin{equation}
\footnotesize
    \mathcal{L}_\text{rec} = \frac{1}{N} \sum_{i=1}^N \left(\eta_1 \lVert \hat{b}_i-b_i\rVert_1 + \eta_2 CE(\hat{\theta}_i,\theta_i) + \eta_3 \lVert\hat{e}_i-e_i\rVert_1  \right)
    \label{eq:rec_loss}
\end{equation}
where $\hat{b}_i, b_i$ are the predicted and ground truth bounding box parameters respectively, $\hat{\theta}_i, \theta_i$ the predicted and ground truth yaw angle of the bounding box and $\hat{e}_i$ the predicted shape code from our model, while $e_i$ is the shape code provided by AtlasNet. We choose $\eta_1=0.4$, $\eta_2=0.2$, $\eta_3=0.4$.

During fine-tuning, we use the following loss for nodes $i\in N$ and edges $j\in M$
\begin{equation}
\footnotesize
    \mathcal{L}_\text{SG} = \frac{1}{N} \sum_{i=1}^N \lambda_1 CE\left(\hat{o}_i,o_i\right) + \frac{1}{M} \sum_{j=1}^M  \lambda_2 BCE\left(\hat{p}_j,p_j\right)
    \label{eq:sg_loss}
\end{equation}
where $\hat{o}_i, o_i$ are the predicted and ground truth object node classes and $\hat{p}_j, p_j$ are the predicted and ground truth predicates for edge $j$. We choose $\lambda_1=0.1$ and $\lambda_2=1.0$.

To deal with class imbalance during fine-tuning, we use a focal loss for both loss terms
\begin{equation}
\footnotesize
    \mathcal{L} = -\alpha_t(1-p_t)^{\gamma}\log p_t
\end{equation}
with $\alpha = 0.25$ and $\gamma = 2$. However, we do not use manual class weighting based on object and predicate occurrences like SGFN~[55].

\section{Dataset Details}
\label{sec:dataset_details}
Following Wald~\etal~[49], our method operates on scene splits with 4-9 objects instead of taking the full 3D scene as input. 
For comparability, we use the exact same pre-split scenes published by Wald~\etal~[49]. 
A full list of the 160 objects and 26 predicates used for the evaluation is in the authors' repository under subset data\footnote{\url{https://github.com/3DSSG/3DSSG.github.io/}}.

For the ScanNet~[12] and S3DIS~[3] pre-training, we use the most updated versions of each dataset available at the time of publishing this paper.
For uniform training samples, we also generate scene splits for the additional datasets to emulate the 3DSSG dataset [49] as best as possible. Moreover, before feeding object point clouds and pairs of objects' point clouds into the PointNets~[41], we apply farthest-point-sampling to downsample the point clouds of each object to at most 1000 points.

\section{Architecture Ablations}
\label{sec:skip_connection}
In \cref{tab:ablations} we provide ablations examining the design choices for our pre-training method. We present results for: Our method without pre-training; Our pre-training only using the shape-loss reconstruction term; Our pre-training only using the bounding box reconstruction term. Further, we provide architecture ablations for our method without utilizing a GCN and without using our proposed skip-connection from Sec. 3. 

We observe that only using the shape-loss during pre-training greatly improves object prediction performance. However, the impact on predicate prediction is small. Only using the bounding box loss term during pre-training improves objects and predicates alike, but the results are worse than using the shape-loss and bounding box-loss together.
A fundamental aspect of our method is the introduced GCN. Without a GCN as a backbone, we observe that our pre-training becomes considerably less effective in the learning of predicates.
In our architecture, we further introduce a singular skip-connection in Eq. 6. This skip-connection serves the role of conserving more context features from the encoder. This skip-connection improves the reconstruction loss and consequently pre-training effectiveness for the entire encoder. 

\begin{table}[t]
\tabcolsep=1.8mm
\centering
\small
\begin{tabular}{@{}lcccc@{}}
\toprule
       & \multicolumn{2}{c}{Object} & \multicolumn{2}{c}{Predicate}\\
      Method & R@5                 & mR@5                & R@3                              & mR@3                              \\
      \midrule
      Ours (w/o pre-train) & 0.63   & 0.30   & 0.94   & 0.57 \\
      \midrule
      Ours (shape-loss only) & 0.77 & 0.39 & 0.94 & 0.49 \\
      Ours (box-loss only) &  0.76 & 0.35 & 0.96 & 0.59 \\
      \midrule
      Ours (w/o GCN) & 0.75 & 0.31 & 0.94 & 0.48\\
      Ours (w/o skip connection) & 0.77 & 0.40 & 0.96 & 0.60\\
\bottomrule
\end{tabular}
\caption{\textbf{Ablations point cloud pre-training approaches}.}
\label{tab:ablations}
\end{table}

\section{Qualitative effect of pre-training}
\label{sec:direct_comp}
 in \cref{fig:graphs_example_compare}, we provide a direct comparison between our method pre-trained using our pre-training approach and the same method trained from scratch on 3D scene graph prediction. We observe that using our pre-training drastically improves the prediction accuracy of nodes and edges. The predicted 3D scene graph using our pre-training is near perfect except for two misclassified objects, while the method trained from scratch predicts many incorrect predicates and also misclassified objects.
 
\section{Scene Graph Results}
\label{sec:sg_resuls}
In \cref{fig:add_sgs_}, we provide additional scene graph visualizations. 
We observe that our network is able to produce almost perfect scene graphs in very diverse scenes. Some common misclassification cases include incorrect edges where either the ground truth is none and our network predicts a relationship or our network does not predict a relationship when it is present.

\begin{figure*}
    \centering
    \begin{subfigure}{0.49\linewidth}
    \centering
    \includegraphics[width=\linewidth,height=0.25\textheight,keepaspectratio]{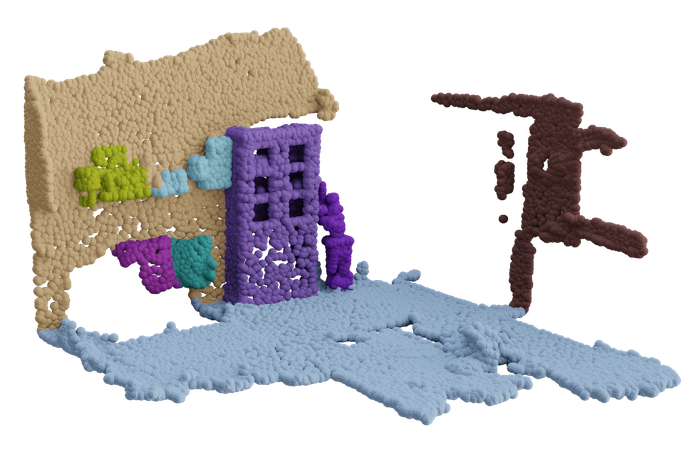}
    \includegraphics[width=\linewidth,height=0.2\textheight,keepaspectratio]{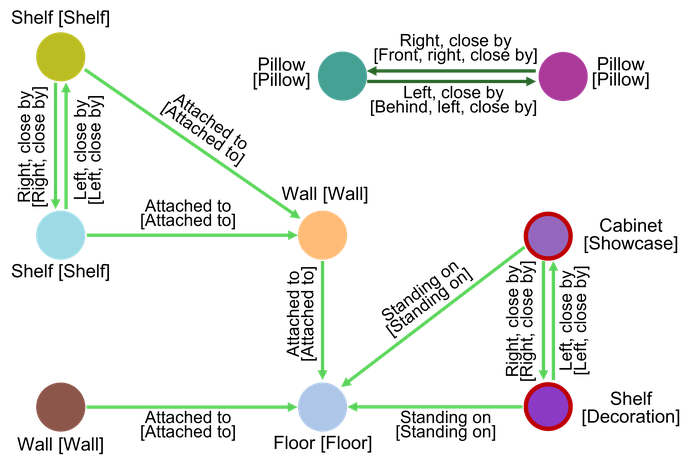}
    \caption{}
    \label{fig:sgs-a}
    \end{subfigure}
    \hfill
    \begin{subfigure}{0.49\linewidth}
    \centering
    \includegraphics[width=\linewidth,height=0.25\textheight,keepaspectratio]{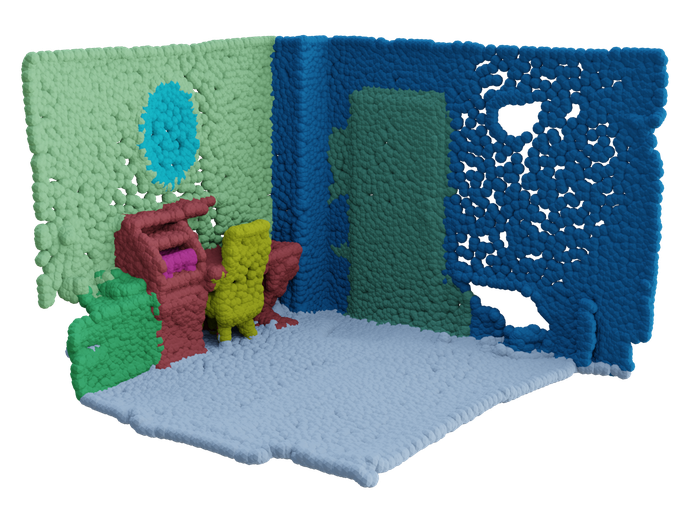}
    \includegraphics[width=\linewidth,height=0.2\textheight,keepaspectratio]{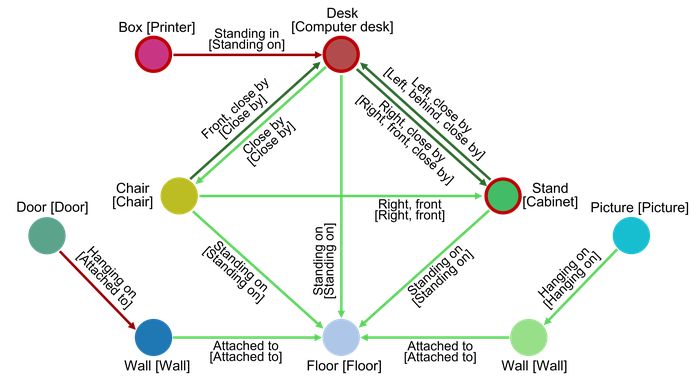}
    \caption{}
    \label{fig:sgs-b}
    \end{subfigure}
    \vspace{1cm}
    
    \begin{subfigure}{0.49\linewidth}
    \centering
    \includegraphics[width=\linewidth,height=0.25\textheight,keepaspectratio]{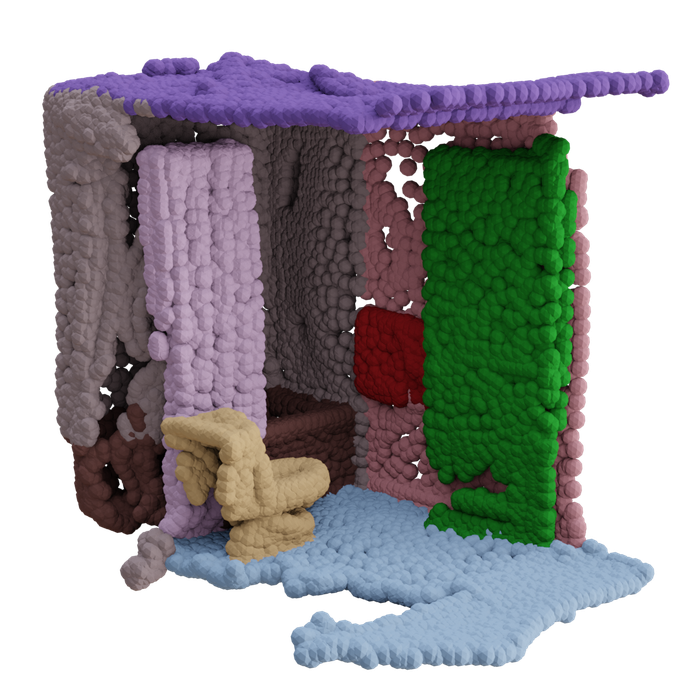}
    \includegraphics[width=\linewidth,height=0.2\textheight,keepaspectratio]{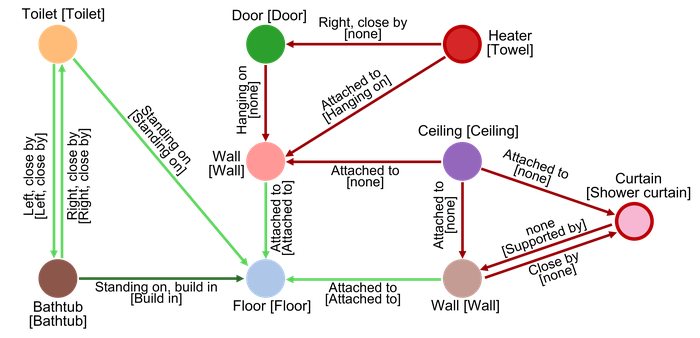}
    \caption{}
    \label{fig:sgs-c}
    \end{subfigure}
    \hfill
    \begin{subfigure}{0.49\linewidth}
    \centering
    \includegraphics[width=\linewidth,height=0.25\textheight,keepaspectratio]{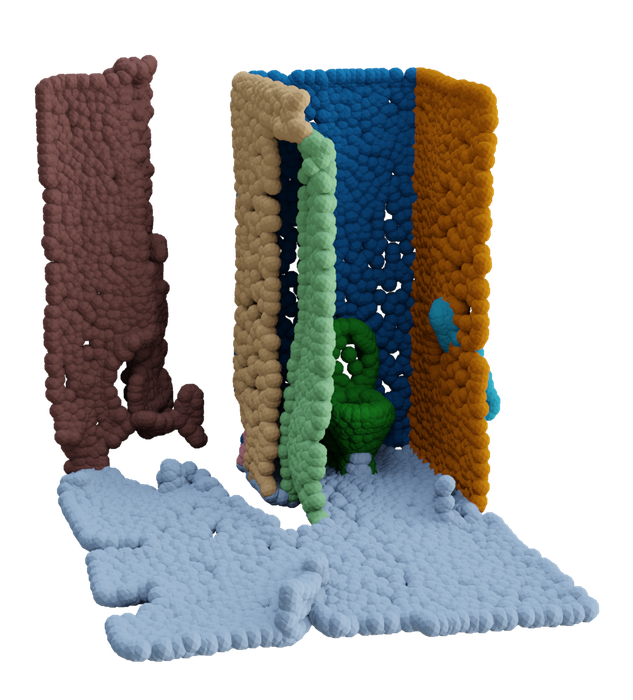}
    \includegraphics[width=\linewidth,height=0.2\textheight,keepaspectratio]{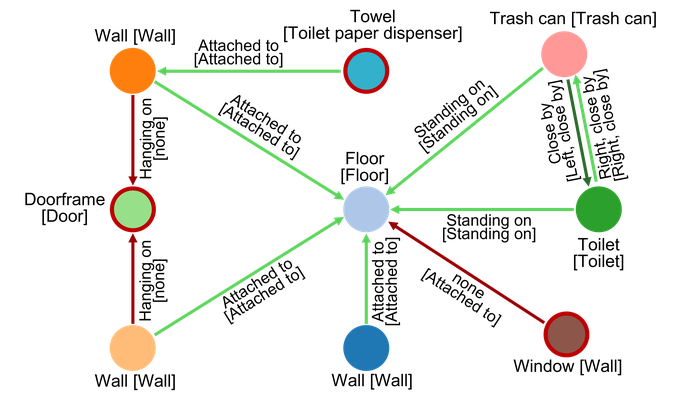}
    \caption{}
    \label{fig:sgs-d}
    \end{subfigure}
\end{figure*}%
\begin{figure*}[ht]\ContinuedFloat
    \begin{subfigure}{0.49\linewidth}
    \centering
    \includegraphics[width=\linewidth,height=0.24\textheight,keepaspectratio]{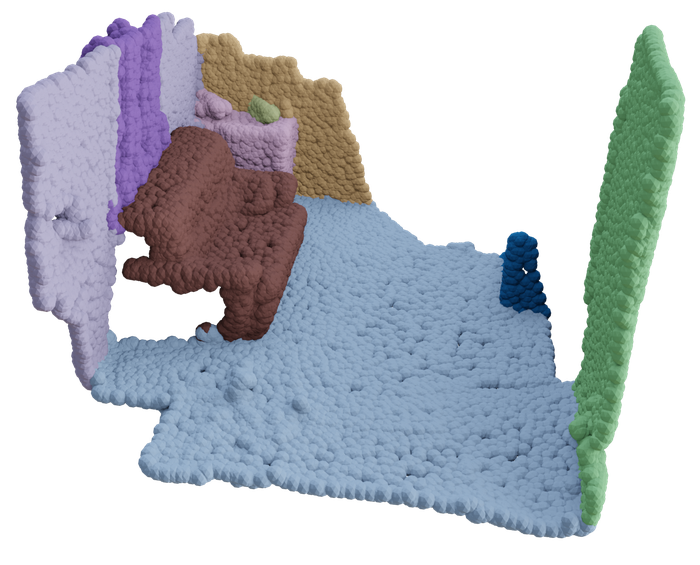}
    \includegraphics[width=\linewidth,height=0.2\textheight,keepaspectratio]{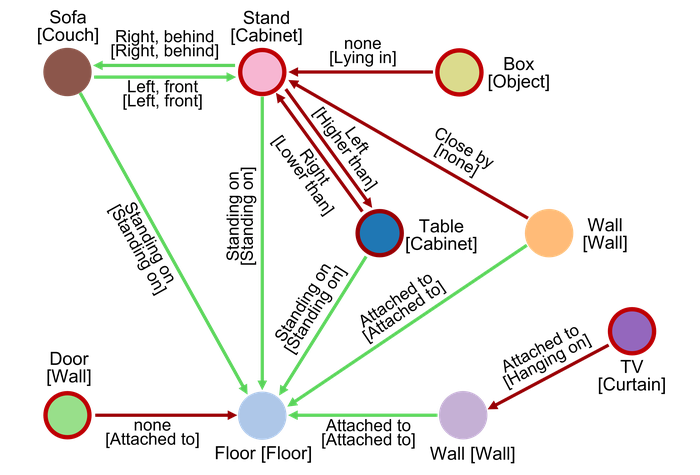}
    \caption{}
    \label{fig:sgs-e}
    \end{subfigure}
    \hfill
    \begin{subfigure}{0.49\linewidth}
    \centering
    \includegraphics[width=\linewidth,height=0.24\textheight,keepaspectratio]{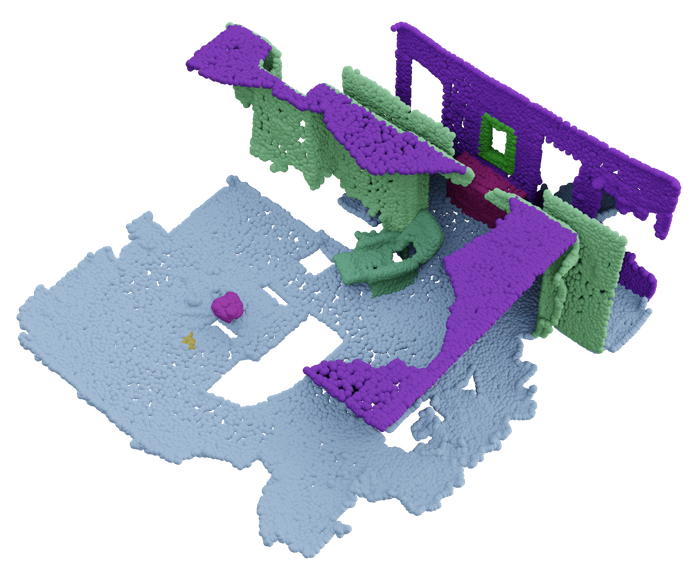}
    \includegraphics[width=\linewidth,height=0.2\textheight,keepaspectratio]{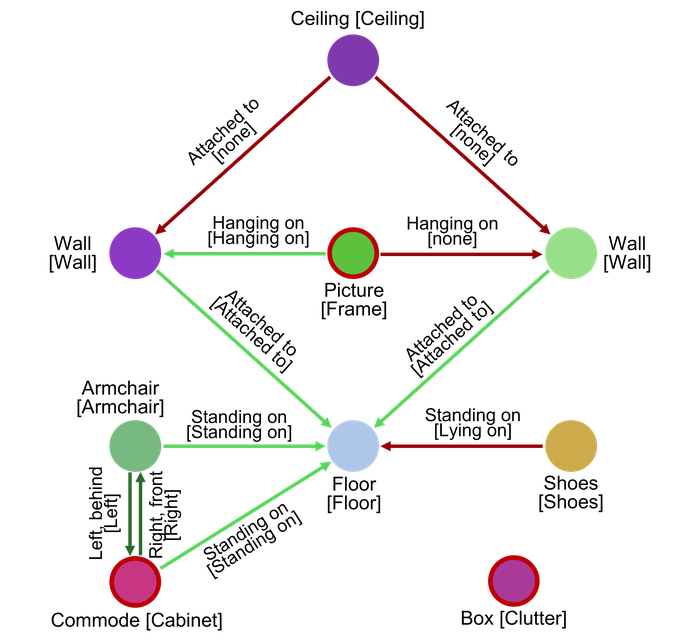}
    \caption{}
    \label{fig:sgs-f}
    \end{subfigure}
    \vspace{1cm}
    
    \begin{subfigure}{0.49\linewidth}
    \centering
    \includegraphics[width=\linewidth,height=0.24\textheight,keepaspectratio]{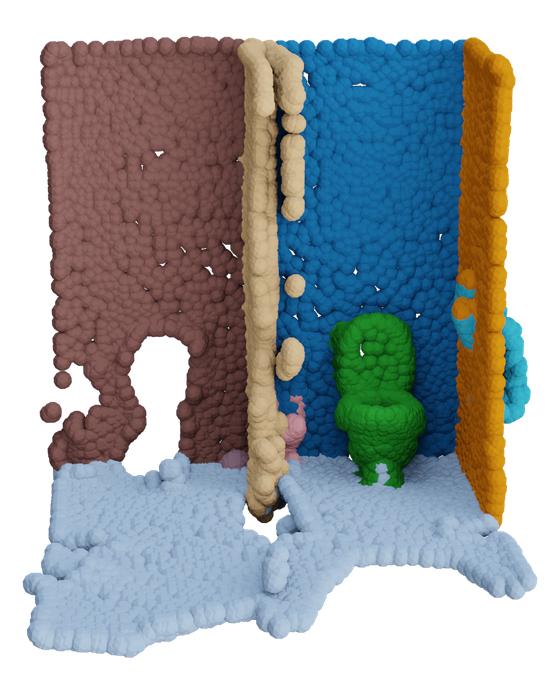}
    \includegraphics[width=\linewidth,height=0.2\textheight,keepaspectratio]{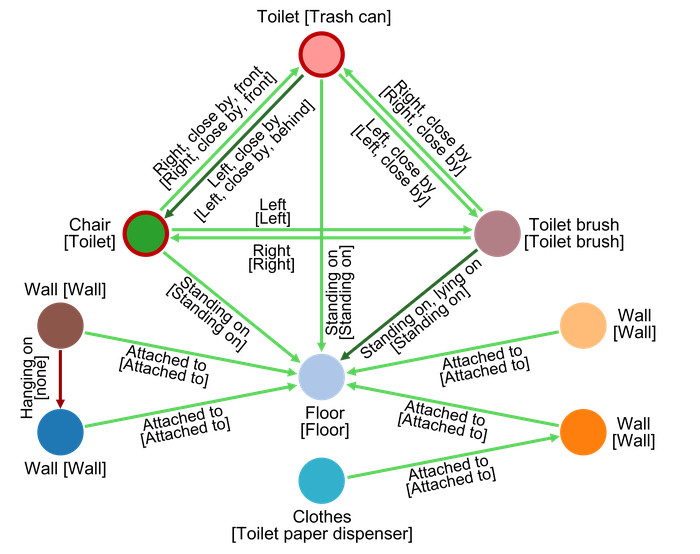}
    \caption{}
    \label{fig:sgs-g}
    \end{subfigure}
    \hfill
    \begin{subfigure}{0.49\linewidth}
    \centering
    \includegraphics[width=\linewidth,height=0.24\textheight,keepaspectratio]{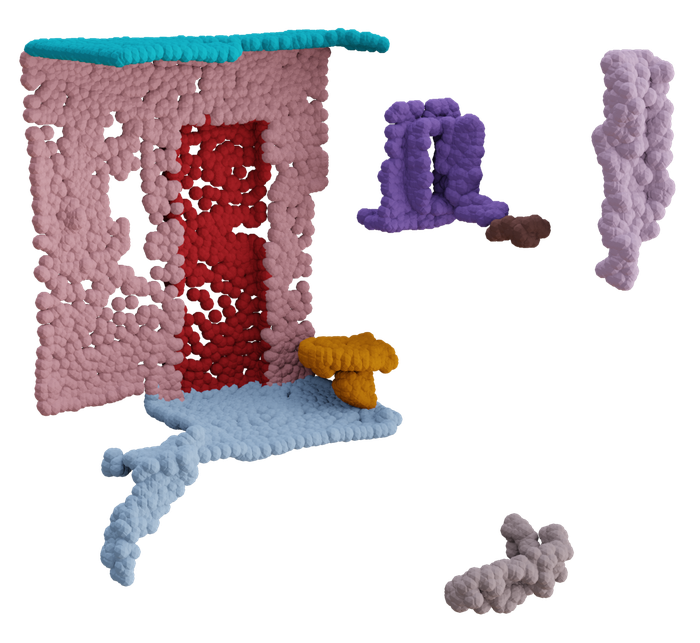}
    \vspace*{1.5cm}
    \includegraphics[width=\linewidth,height=0.2\textheight,keepaspectratio]{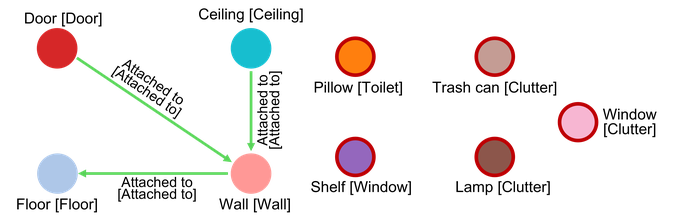}
    \caption{}
    \label{fig:sgs-h}
    \end{subfigure}
    \caption{\textbf{Qualitative scene graph results from \ours.} Top: 3D scene; Bottom: Predicted 3D scene graph.%
    \label{fig:add_sgs_}}
\end{figure*}

\section{Scene Generation Results}
\label{sec:scene_results}
In \cref{fig:add_scen_gens}, we provide additional scene reconstructions. The shown scene generations support those shown in the paper. Although the generated shapes are not perfect, our model seems to preserve the relationships in the original scenes. 

\begin{figure*}
    \centering
    \begin{subfigure}{0.48\linewidth}
    \includegraphics[width=\linewidth]{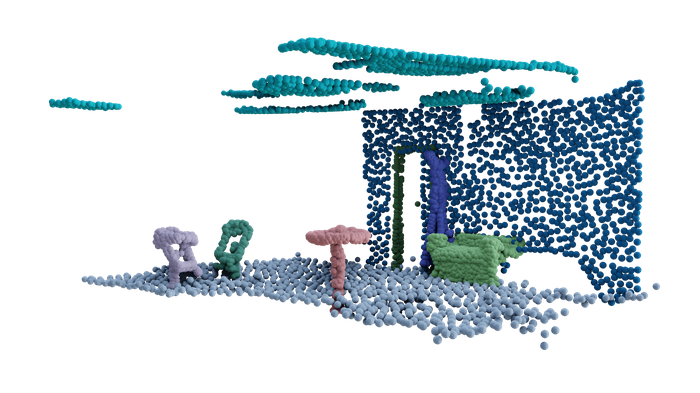}
    \includegraphics[width=\linewidth]{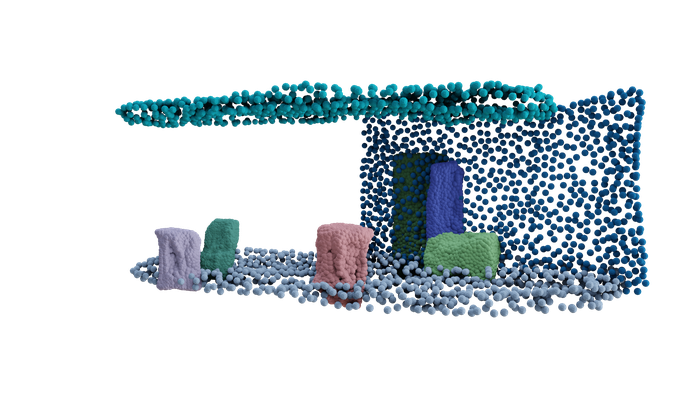}
    \caption{}
    \label{fig:scene_gens-a}
    \end{subfigure}
    \hfill
    \begin{subfigure}{0.48\linewidth}
    \includegraphics[width=\linewidth]{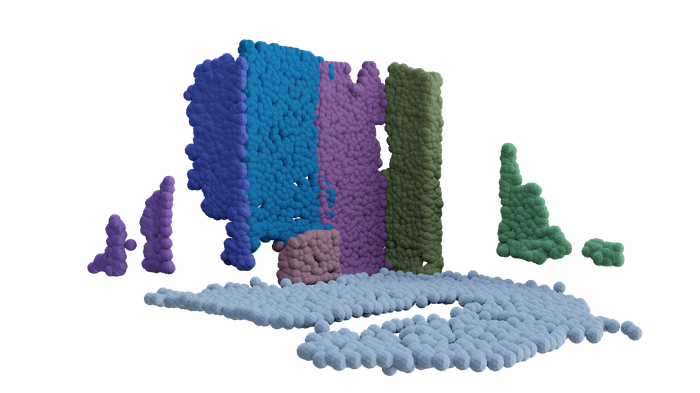}
    \includegraphics[width=\linewidth]{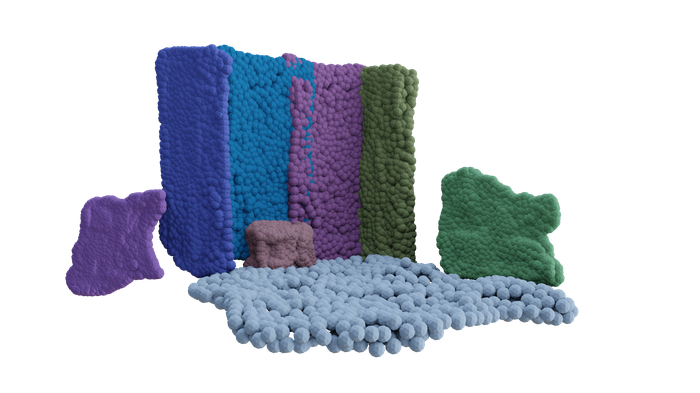}
    \caption{}
    \label{fig:scene_gens-b}
    \end{subfigure}
    \vspace{1cm}
    
    \begin{subfigure}{0.48\linewidth}
    \includegraphics[width=\linewidth]{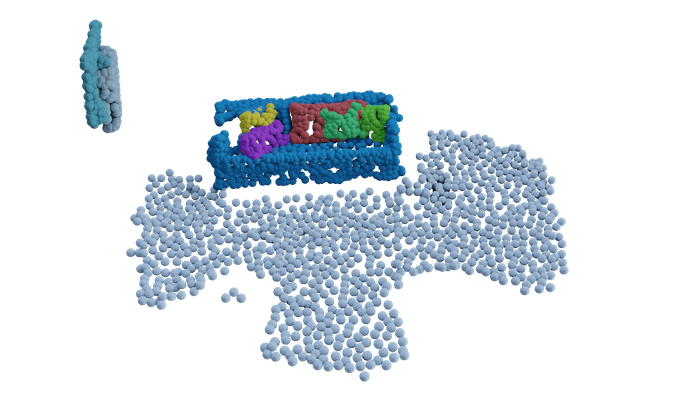}
    \includegraphics[width=\linewidth]{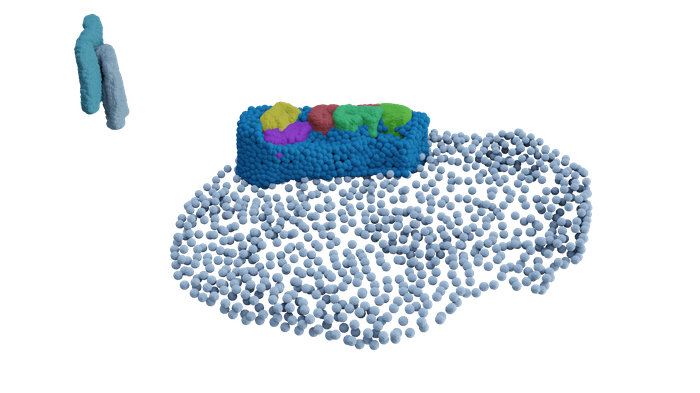}
    \caption{}
    \label{fig:scene_gens-c}
    \end{subfigure}
    \hfill
    \begin{subfigure}{0.48\linewidth}
    \includegraphics[width=\linewidth]{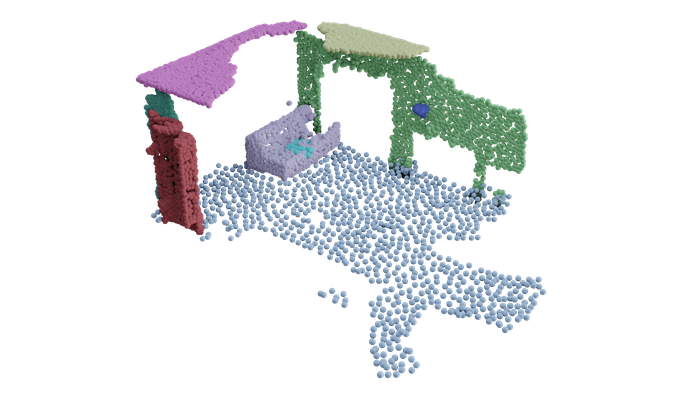}
    \includegraphics[width=\linewidth]{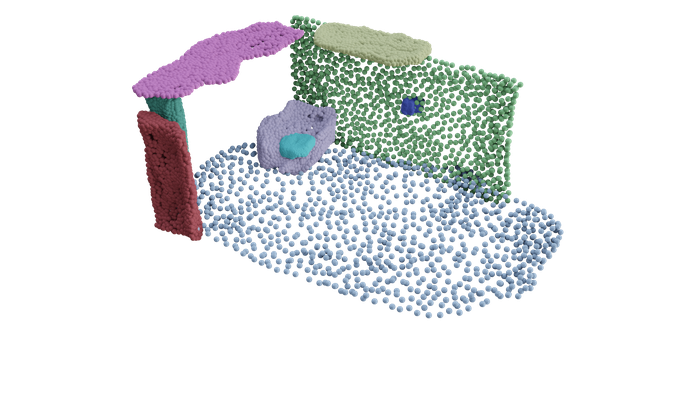}
    \caption{}
    \label{fig:scene_gens-d}
    \end{subfigure}
    
\end{figure*}%
\begin{figure*}[ht]\ContinuedFloat
    \begin{subfigure}{0.48\linewidth}
    \includegraphics[width=\linewidth]{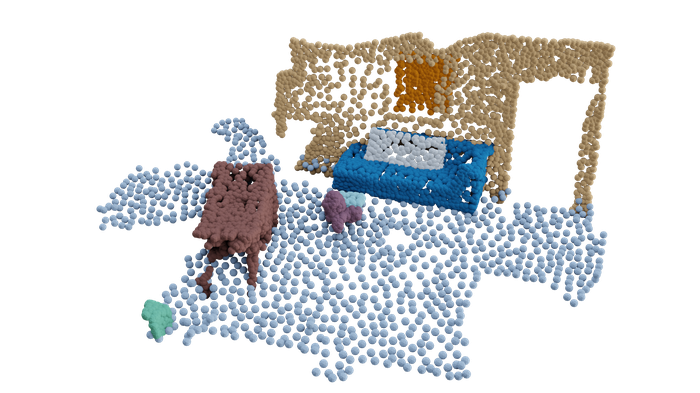}
    \includegraphics[width=\linewidth]{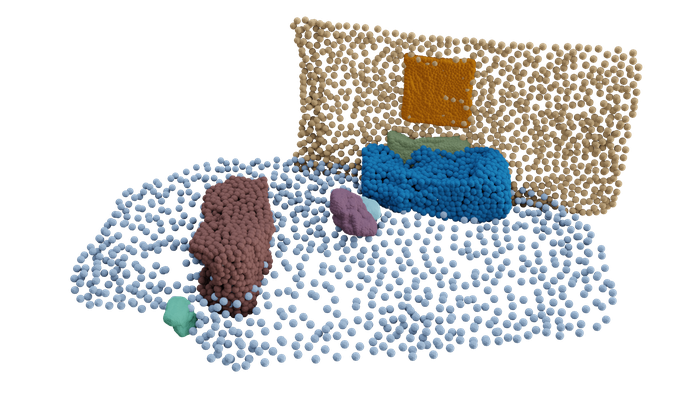}
    \caption{}
    \label{fig:scene_gens-e}
    \end{subfigure}
    \hfill
    \begin{subfigure}{0.48\linewidth}
    \includegraphics[width=\linewidth]{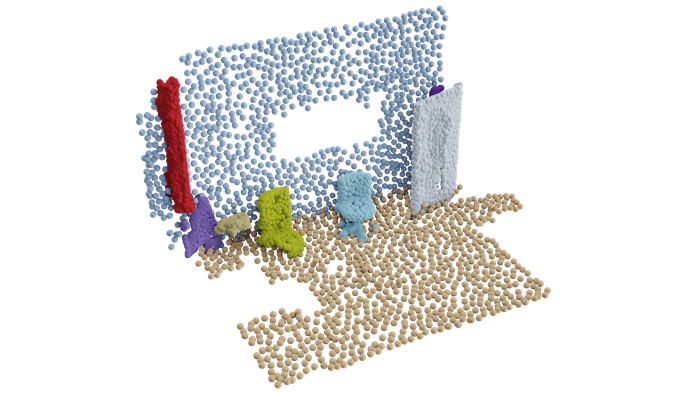}
    \includegraphics[width=\linewidth]{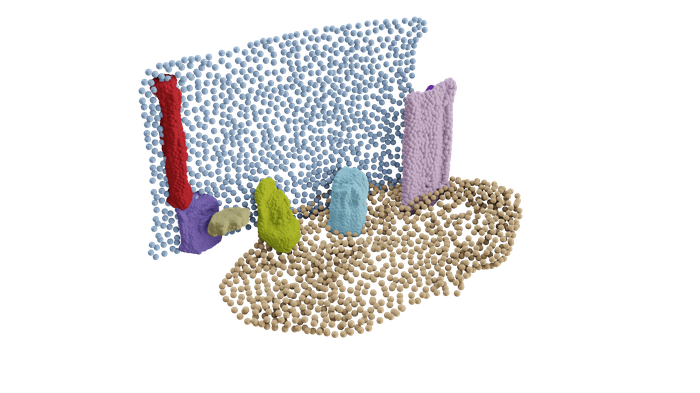}
    \caption{}
    \label{fig:scene_gens-f}
    \end{subfigure}
    
    \vspace{1cm}
    \begin{subfigure}{0.48\linewidth}
    \includegraphics[width=\linewidth]{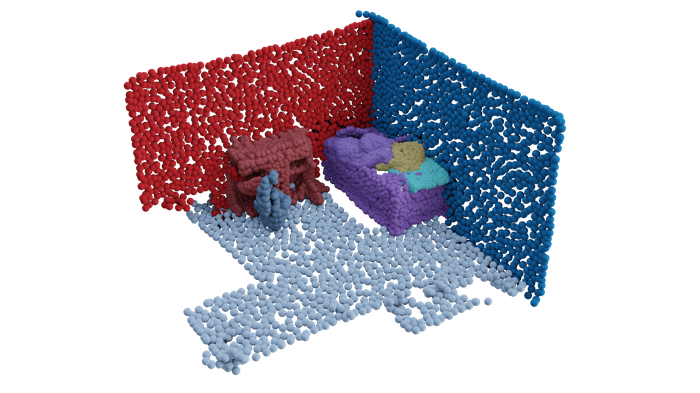}
    \includegraphics[width=\linewidth]{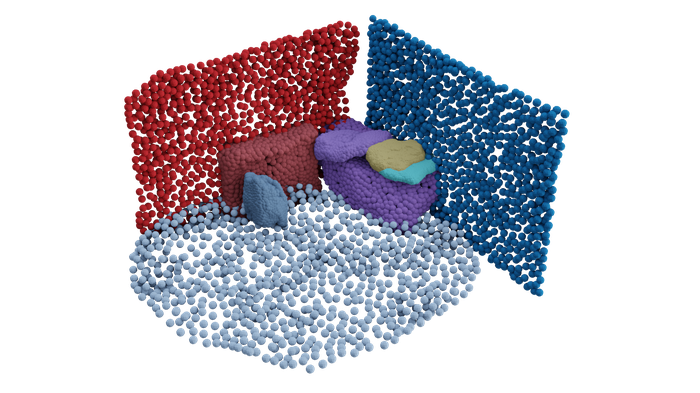}
    \caption{}
    \label{fig:scene_gens-g}
    \end{subfigure}
    \hfill
    \begin{subfigure}{0.48\linewidth}
    \includegraphics[width=\linewidth]{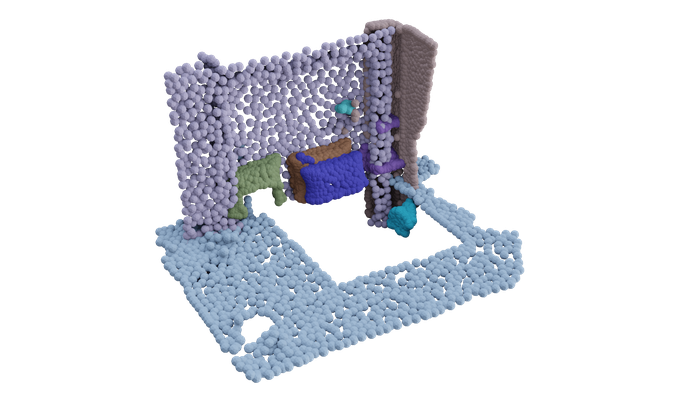}
    \includegraphics[width=\linewidth]{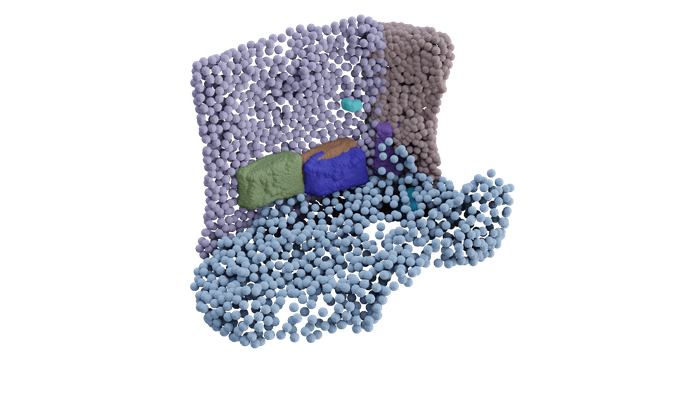}
    \caption{}
    \label{fig:scene_gens-h}
    \end{subfigure}
    \caption{\textbf{Qualitative scene generation results from \ours.} Top: Original 3D scene; Bottom: Reconstructed scene using \ours.}
    \label{fig:add_scen_gens}
\end{figure*}